\documentclass[preprint,12pt]{elsarticle}

\usepackage{amsmath,amssymb,amsfonts}
\usepackage{graphicx}
\usepackage{xcolor}
\usepackage{booktabs}
\usepackage{multirow}
\usepackage{hyperref}
\usepackage{float}
\usepackage{algorithm}
\usepackage{algpseudocode}
\usepackage{tikz}
\usetikzlibrary{arrows.meta,positioning,calc}
\usepackage{pgfplots}
\pgfplotsset{compat=1.18}
\usepgfplotslibrary{fillbetween}
\usepackage{comment}
\usepackage{natbib}

\newcommand{\metricup}{$\uparrow$}
\newcommand{\metricdown}{$\downarrow$}

\begin{document}

\begin{frontmatter}

\title{Statistical Roughness-Informed Machine Unlearning }
\author[inst1]{Mohammad Partohaghighi}
\author[inst2]{Roummel Marcia}
\author[inst3]{Bruce J. West}
\author[inst4]{YangQuan Chen}

\address[inst1]{Electrical Engineering and Computer Science, University of California, Merced, CA 95343, USA}
\address[inst2]{Department of Applied Mathematics, University of California, Merced, CA 95343, USA}
\address[inst3]{Department of Innovation and Research, North Carolina State University, Raleigh, NC, USA}
\address[inst4]{Mechatronics, Embedded Systems and Automation (MESA) Lab, Department of Mechanical Engineering, School of Engineering, University of California, Merced, CA 95343, USA}

\begin{abstract}
Machine unlearning aims to remove the influence of a designated forget set from a trained model while preserving utility on the retained data.
In modern deep networks, approximate unlearning frequently fails under large or adversarial deletions due to pronounced layer-wise heterogeneity: some layers exhibit stable, well-regularized representations while others are brittle, undertrained, or overfit, so naive update allocation can trigger catastrophic forgetting or unstable dynamics.
We propose Statistical-Roughness Adaptive Gradient Unlearning (SRAGU), a mechanism-first unlearning algorithm that reallocates unlearning updates using layer-wise statistical roughness operationalized via heavy-tailed spectral diagnostics of layer weight matrices.
Starting from an Adaptive Gradient Unlearning (AGU) sensitivity signal computed on the forget set, SRAGU estimates a WeightWatcher-style heavy-tailed exponent for each layer, maps it to a bounded spectral stability weight, and uses this stability signal to spectrally reweight the AGU sensitivities before applying the same minibatch update form.
This concentrates unlearning motion in spectrally stable layers while damping updates in unstable or overfit layers, improving stability under hard deletions.
We evaluate unlearning via behavioral alignment to a gold retrained reference model trained from scratch on the retained data, using empirical prediction-divergence and KL-to-gold proxies on a forget-focused query set; we additionally report membership inference auditing as a complementary leakage signal, treating forget-set points as should-be-forgotten members during evaluation.
\end{abstract}


\end{frontmatter}
\section{Introduction}
\label{sec:introduction}

\paragraph{Context and Motivation}
Modern machine learning systems operate under continuous change: new data arrives, policies evolve, deployments are audited, and users or institutions may request that specific records no longer influence a model.
This reality is reinforced by regulatory frameworks that formalize deletion and control rights over personal data, elevating \emph{machine unlearning} from a niche security curiosity to a practical requirement in the model lifecycle \citep{gdpr2016,ccpa_oag}.
At a high level, unlearning seeks to remove the influence of designated training records while preserving utility on retained data, approximating the behavior of a model trained without the removed information \citep{cao2015unlearning,ginart2019forgetyou,bourtoule2021unlearning}.

\paragraph{Problem and Challenge}
The ideal solution is conceptually simple: retrain from scratch on the retained data.
In realistic settings, however, retraining is often prohibitively expensive due to repeated deletion requests, large-scale models, and production constraints.
This gap has motivated a spectrum of efficient unlearning techniques, but efficiency alone is not enough: approximate unlearning can be unstable, degrade utility, or provide weak deletion fidelity under adversarial or high-impact forget requests \citep{guo2020certified,sekhari2021remember,kurmanji2023unbounded,zhao2024hard,jia2023sparsity}.
Moreover, unlearning is tightly entangled with the broader problem of model leakage and memorization, where attackers may infer membership or extract training content, making robust removal mechanisms particularly important in sensitive domains \citep{shokri2017membership,carlini2021extracting}.

\paragraph{Prior Work Positioning}
Early formulations framed unlearning as a system-level capability and explored transformations that support data removal \citep{cao2015unlearning}.
Subsequent work developed deletion-efficient learning principles and scalable training frameworks such as sharding-and-slicing to enable partial retraining \citep{ginart2019forgetyou,bourtoule2021unlearning}.
For deep networks, practical approaches include selective forgetting and weight scrubbing \citep{golatkar2020eternal,golatkar2020outside}, update-caching and rollback-style strategies \citep{graves2021amnesiac}, decision-boundary shifting \citep{chen2023boundary}, and saliency-guided unlearning updates \citep{fan2024salun}.
Recent research has also sharpened the understanding of what makes unlearning easy or hard, and proposed mechanisms that scale across unlearning desiderata and threat models \citep{kurmanji2023unbounded,zhao2024hard,jia2023sparsity}.
Surveys provide structured taxonomies, metrics, and open problems across these families \citep{xu2023survey,nguyen2022survey,shaik2023survey,wang2024survey}.

\paragraph{Key Gap: Layer-wise Heterogeneity and Optimization Geometry.}
A recurring practical obstacle is that deep networks are not uniform objects: layers can differ substantially in conditioning, stability, and how brittle their representations are under targeted updates.
This heterogeneity interacts with optimization geometry, where sharpness, flatness, and basin structure have long been connected to stability and generalization behavior \citep{keskar2017largebatch,li2018losslandscape,hochreiter1997flat}.
In the unlearning setting, an update rule that is globally sensible can still be locally destructive when it concentrates changes in fragile regions of the network.
This motivates unlearning mechanisms that are not only sensitive to \emph{which parameters matter} for forgetting, but also aware of \emph{which layers can safely absorb} the required modifications.

\paragraph{Our Idea: SRAGU}
This paper introduces Statistical-Roughness Adaptive Gradient Unlearning (SRAGU), a drop-in enhancement to the Adaptive Gradient Unlearning (AGU) baseline \citep{ghannam2025agu}.
AGU prioritizes unlearning updates using sensitivity signals computed from the forget set, offering an efficient and appealing starting point.
SRAGU retains this sensitivity-driven prioritization but modulates it using a layer-wise stability signal derived from heavy-tailed spectral diagnostics of weight matrices, inspired by empirical heavy-tailed phenomena in trained deep networks and their interpretation through random-matrix perspectives \citep{martin2019heavytailed,martin2021implicit}.
Intuitively, SRAGU steers unlearning effort toward layers that appear spectrally stable and away from layers that exhibit signatures of brittleness or overfitting, improving robustness without adding expensive per-example instrumentation beyond the baseline workflow.
This direction is also aligned with the broader trend of expanding unlearning beyond standard classification into modern generative and foundation-model settings, where targeted removal and behavior editing have become increasingly relevant \citep{gandikota2023erasing,isonuma2024untrac}.

\paragraph{Contributions}
We make the following contributions.
First, we identify a concrete failure mode of sensitivity-driven unlearning that arises from layer-wise heterogeneity and optimization geometry, clarifying why efficient forgetting can become unstable.
Second, we propose SRAGU, a stability-aware extension of AGU that uses heavy-tailed spectral diagnostics of layer weights to modulate unlearning updates.
Third, we provide an implementation-ready methodology that integrates these diagnostics into unlearning with minimal disruption to standard training and evaluation pipelines.
Finally, we position SRAGU within the broader unlearning landscape and connect its mechanism to prior efficient unlearning paradigms, modern desiderata and hardness considerations, and known spectral/optimization phenomena \citep{xu2023survey,nguyen2022survey,wang2024survey,zhao2024hard}.

\paragraph{Roadmap}
The remainder of the paper is organized as follows.
Section~\ref{sec:methodology} reviews machine unlearning and the AGU baseline in the context of representative unlearning families and evaluation desiderata.
Section~\ref{subsec:sragu} presents SRAGU and  and Section~\ref{sec:mechanism} presents mechanistic analysis.
Section~\ref{sec:experiments} provides an empirical evaluation across deletion settings and baselines, and Section~\ref{sec:conclusion} concludes with limitations and future directions.

\section{Methodology}
\label{sec:methodology}
The Roughness-informed machine unlearning framework enhances the Adaptive Gradient Unlearning (AGU) baseline by explicitly modeling geometric roughness in the loss surface and statistical roughness in layer-wise parameter distributions, enabling stable and reliable forgetting.
This section formalizes the problem, reviews the AGU baseline, details its limitations, and presents the proposed SRAGU algorithms.

\subsection{Problem Formulation}
Assume a pre-trained model with parameters \(\boldsymbol{\phi} \in \mathbb{R}^n\), a forget subset \(\mathcal{D}_{\text{forget}} = \{(u_i, v_i)\}_{i=1}^{|\mathcal{D}_{\text{forget}}|}\), and a retained data collection.
The objective is to derive an adjusted parameter configuration \(\boldsymbol{\phi}'\) such that the model \(F(\cdot; \boldsymbol{\phi}')\) behaves as if trained exclusively on the retained data, i.e., without influence from \(\mathcal{D}_{\text{forget}}\).

We quantify unlearning leakage via empirical alignment between the unlearned model and a gold retrained model (trained from scratch without \(\mathcal{D}_{\text{forget}}\)), following the prediction-divergence protocol used by AGU \citep{ghannam2025agu}.
We define the empirical leakage proxy as:
\begin{equation}
\epsilon_{\text{pred}} =
\max_{(x,y)\in \mathcal{D}_{\text{forget}}}
\left\| p_{\text{unlearn}}(\cdot \mid x) - p_{\text{retrain}}(\cdot \mid x)\right\|_{1},
\label{eq:eps_pred}
\end{equation}
where \(p(\cdot\mid x)\) is the softmax predictive distribution. Smaller \(\epsilon_{\text{pred}}\) indicates stronger forgetting (closer behavior to exact retraining) and reduced residual influence.

The process must also preserve performance on retained samples, quantified through accuracy or loss.
The challenge lies in approximating ideal unlearning equivalent to complete retraining efficiently and robustly, particularly under large deletion ratios or adversarially chosen forget sets.

\subsection{Baseline: Adaptive Gradient Unlearning (AGU)}
\label{subsec:agu}

AGU~\citep{ghannam2025agu} is an efficient approximate unlearning approach that prioritizes parameter updates according to responsiveness to the forget set $\mathcal{D}_{\mathrm{forget}}$.
Let $\boldsymbol{\phi}_0\in\mathbb{R}^n$ denote the pre-unlearning parameters and let $\phi_j$ be the $j$-th scalar parameter.

\paragraph{Sensitivity indicator.}
For each parameter $\phi_j$, AGU computes a sensitivity score as the mean squared gradient magnitude over the forget set, evaluated at $\boldsymbol{\phi}_0$:
\begin{equation}
R_j
=
\frac{1}{|\mathcal{D}_{\mathrm{forget}}|}
\sum_{(u_i,v_i)\in \mathcal{D}_{\mathrm{forget}}}
\left(
\nabla_{\phi_j}\,\mathcal{L}(u_i,v_i;\boldsymbol{\phi}_0)
\right)^2,
\label{eq:agu_sensitivity}
\end{equation}
where $\mathcal{L}(\cdot,\cdot;\boldsymbol{\phi})$ is the per-example loss (e.g., cross-entropy), and $\nabla_{\phi_j}$ denotes the partial derivative w.r.t.\ parameter $\phi_j$.

\paragraph{Normalization.}
To obtain bounded weights comparable across parameters, AGU rescales:
\begin{equation}
\bar{R}_j
=
\frac{R_j}{\max_{k\in\{1,\dots,n\}} R_k +\epsilon_R},
\label{eq:agu_normalized_sensitivity}
\end{equation}
where $\epsilon_R>0$ is a small numerical stabilizer (e.g., $\epsilon_R=10^{-12}$), ensuring well-defined normalization even when $\max_k R_k$ is extremely small.
By construction, $\bar{R}_j\in[0,1]$ up to numerical precision.

\paragraph{Unlearning update.}
Given a minibatch $\mathcal{B}\subset \mathcal{D}_{\mathrm{forget}}$, AGU applies the sensitivity-weighted update:
\begin{equation}
\phi_j
\leftarrow
\phi_j
-
\alpha\,\bar{R}_j\,
\nabla_{\phi_j}\mathcal{L}(\mathcal{B};\boldsymbol{\phi}),
\label{eq:agu_update}
\end{equation}
where $\alpha>0$ is the unlearning learning rate, and $\mathcal{L}(\mathcal{B};\boldsymbol{\phi})$ is the minibatch-averaged loss over $\mathcal{B}$.

\paragraph{Stopping rule .}
In our paper (Setting A), the behavioral metrics $\epsilon_{\mathrm{pred}}$ and $D_{\mathrm{KL}}$ are defined as distances to the gold retraining model ORTR; hence they are \emph{exactly zero} for ORTR by definition. Because computing $\epsilon_{\mathrm{pred}}(\boldsymbol{\phi})$ requires ORTR outputs, it is primarily an \emph{evaluation} metric and is not required for the practical execution of AGU.
Accordingly, we use an implementation-ready stopping rule that does not require ORTR:
\begin{equation}
\|\boldsymbol{\phi}-\boldsymbol{\phi}_0\|_2 < \kappa
\quad \textbf{or} \quad
t\ge T,
\label{eq:agu_stop}
\end{equation}
where $\kappa>0$ is a drift tolerance and $T$ is a maximum number of unlearning steps/epochs.
(Optional oracle variant, analysis only): when ORTR predictions are available for the same seed and deletion setting, one may additionally stop if $\epsilon_{\mathrm{pred}}(\boldsymbol{\phi})\le \epsilon_{\mathrm{target}}$, but we treat this as an oracle diagnostic rather than a requirement of AGU.

\begin{algorithm}[H]
\caption{Adaptive Gradient Unlearning (AGU)}
\label{alg:agu}
\begin{algorithmic}[1]
\Require Pre-unlearning parameters $\boldsymbol{\phi}_0$, forget set $\mathcal{D}_{\mathrm{forget}}$, learning rate $\alpha$, drift tolerance $\kappa$, max steps $T$, stabilizer $\epsilon_R$
\Ensure Unlearned parameters $\boldsymbol{\phi}$
\State Initialize $\boldsymbol{\phi} \leftarrow \boldsymbol{\phi}_0$
\State Compute sensitivities $\{R_j\}_{j=1}^n$ using Eq.~\eqref{eq:agu_sensitivity}
\State Compute $R_{\max}\leftarrow \max_k R_k$
\For{$j=1$ to $n$}
    \State $\bar{R}_j \leftarrow \dfrac{R_j}{R_{\max} +\epsilon_R}$ \Comment{Eq.~\eqref{eq:agu_normalized_sensitivity}}
\EndFor
\For{$t=1$ to $T$}
    \For{each minibatch $\mathcal{B}\subset \mathcal{D}_{\mathrm{forget}}$}
        \State Compute $G_{\mathcal{B}} \leftarrow \nabla_{\boldsymbol{\phi}} \mathcal{L}(\mathcal{B};\boldsymbol{\phi})$
        \For{$j=1$ to $n$}
            \State $\phi_j \leftarrow \phi_j - \alpha \,\bar{R}_j\, G_{\mathcal{B}}[j]$ \Comment{Eq.~\eqref{eq:agu_update}}
        \EndFor
    \EndFor
    \If{$\|\boldsymbol{\phi}-\boldsymbol{\phi}_0\|_2<\kappa$}
        \State \textbf{break} \Comment{Eq.~\eqref{eq:agu_stop}}
    \EndIf
\EndFor
\State Compute $\epsilon_{\mathrm{pred}}$ and $D_{\mathrm{KL}}$ \emph{only for evaluation} under Setting A (distances to ORTR).
\Return $\boldsymbol{\phi}$
\end{algorithmic}
\end{algorithm}

\subsubsection{Limitations of AGU}
\label{subsubsec:agu_limits}
Although AGU prioritizes parameters responsive to $\mathcal{D}_{\mathrm{forget}}$, it implicitly assumes that all layers can safely accommodate the induced update magnitudes.
In practice, deep networks exhibit strong layer-wise heterogeneity: some layers are spectrally stable and well-trained, while others are brittle (undertrained) or overfit.
This mismatch can yield unstable unlearning dynamics when sensitive parameters reside in unstable layers.
We address this limitation by incorporating \emph{statistical roughness} as a layer-wise stability modulator in SRAGU.




\section{Statistical-Roughness Adaptive Gradient Unlearning (SRAGU)}
\label{subsec:sragu}

SRAGU improves AGU by reweighting AGU sensitivities using layer-wise statistical roughness derived from heavy-tailed spectral diagnostics.
Throughout, we follow Setting A: $\epsilon_{\mathrm{pred}}$ and $D_{\mathrm{KL}}$ are computed as distances to ORTR outputs, hence they are exactly zero for ORTR by definition and are used for evaluation (not required to run SRAGU).

\paragraph{Spectral diagnostics used by SRAGU.}
For each layer $l\in\{1,\dots,L\}$, let $\mathbf{W}_l\in\mathbb{R}^{m\times n}$ denote the weight matrix of layer $l$, where $m$ and $n$ are its row and column dimensions, respectively.
We form the (uncentered) correlation/Gram matrix
\begin{equation}
\mathbf{C}_l
=
\frac{1}{m}\mathbf{W}_l^\top \mathbf{W}_l .
\label{eq:correlation_matrix}
\end{equation}
Let $\{\lambda_{l,i}\}_{i=1}^{s}$ be the eigenvalues of $\mathbf{C}_l$ with $s=\min(m,n)$.
We fit the top $h=\lfloor \tau s \rfloor$ eigenvalues to a power law
\begin{equation}
p(\lambda)\propto \lambda^{-\xi_l},
\qquad
\lambda\in\{\lambda_{l,1},\dots,\lambda_{l,h}\},
\label{eq:power_law}
\end{equation}
to obtain the heavy-tail exponent $\xi_l$.

\paragraph{Bounded spectral stability weight.}
We map $\xi_l$ to a bounded stability weight $\nu_l\in(0,1)$ using a smooth “in-range” gate
that emphasizes an empirically stable heavy-tailed regime (e.g., $2<\xi_l<4$):
\begin{equation}
\nu_l
=
\sigma\!\big(d_1(4-\xi_l)\big)\;\sigma\!\big(d_2(\xi_l-2)\big),
\qquad
\sigma(z)=\frac{1}{1+e^{-z}} .
\label{eq:spectral_weight}
\end{equation}

\paragraph{Spectrally reweighted sensitivities.}
Let $l(j)$ denote the layer containing parameter $\phi_j$.
SRAGU first computes the AGU sensitivities $R_j$ at $\boldsymbol{\phi}_0$ (Eq.~\eqref{eq:agu_sensitivity}), then applies layer-wise spectral modulation:
\begin{equation}
R'_j = R_j \cdot \nu_{l(j)},
\qquad
\bar{R}^{(\mathrm{SR})}_j
=
\frac{R'_j}{\max_{k\in\{1,\dots,n\}} R'_k +\epsilon_R},
\label{eq:sragu_weighted_sensitivity}
\end{equation}
where the same numerical stabilizer $\epsilon_R>0$ used in AGU is retained for full consistency.
SRAGU then uses the same update form as AGU (Eq.~\eqref{eq:agu_update}) by replacing $\bar{R}_j$ with $\bar{R}^{(\mathrm{SR})}_j$.

\paragraph{Stopping rule (matches AGU, no ORTR required).}
SRAGU uses the same implementation-ready stopping rule as AGU:
\begin{equation}
\|\boldsymbol{\phi}-\boldsymbol{\phi}_0\|_2 < \kappa
\quad \textbf{or} \quad
t\ge T.
\label{eq:sragu_stop}
\end{equation}
(Optional oracle variant, analysis only): if ORTR outputs are available for the same seed/setting, one may also monitor $\epsilon_{\mathrm{pred}}(\boldsymbol{\phi})$ as a diagnostic; this does not change the algorithmic definition.

\begin{algorithm}[H]
\caption{Statistical-Roughness Adaptive Gradient Unlearning (SRAGU)}
\label{alg:sragu}
\begin{algorithmic}[1]
\Require Pre-unlearning parameters $\boldsymbol{\phi}_0$, forget set $\mathcal{D}_{\mathrm{forget}}$, learning rate $\alpha$, drift tolerance $\kappa$, max steps $T$, tail fraction $\tau$, damping constants $d_1,d_2$, stabilizer $\epsilon_R$
\Ensure Unlearned parameters $\boldsymbol{\phi}$
\State Initialize $\boldsymbol{\phi}\leftarrow \boldsymbol{\phi}_0$
\For{$l=1$ to $L$}
    \State Let $\mathbf{W}_l$ be the weight matrix of layer $l$
    \State Compute $\mathbf{C}_l=\frac{1}{m}\mathbf{W}_l^\top \mathbf{W}_l$ \Comment{Eq.~\eqref{eq:correlation_matrix}}
    \State Fit top $h=\lfloor \tau s\rfloor$ eigenvalues ($s=\min(m,n)$) to $p(\lambda)\propto \lambda^{-\xi_l}$ \Comment{Eq.~\eqref{eq:power_law}}
    \State Compute $\nu_l$ from $\xi_l$ \Comment{Eq.~\eqref{eq:spectral_weight}}
\EndFor
\State Compute $R_j$ using Eq.~\eqref{eq:agu_sensitivity} at $\boldsymbol{\phi}_0$
\State Form $R'_j \leftarrow R_j\cdot \nu_{l(j)}$ and $\bar{R}^{(\mathrm{SR})}_j$ using Eq.~\eqref{eq:sragu_weighted_sensitivity}
\For{$t=1$ to $T$}
    \For{each minibatch $\mathcal{B}\subset \mathcal{D}_{\mathrm{forget}}$}
        \State Compute $G_{\mathcal{B}}\leftarrow \nabla_{\boldsymbol{\phi}} \mathcal{L}(\mathcal{B};\boldsymbol{\phi})$
        \For{$j=1$ to $n$}
            \State $\phi_j \leftarrow \phi_j - \alpha \,\bar{R}^{(\mathrm{SR})}_j\, G_{\mathcal{B}}[j]$
            \Comment{Eq.~\eqref{eq:agu_update} with $\bar{R}_j\leftarrow\bar{R}^{(\mathrm{SR})}_j$}
        \EndFor
    \EndFor
    \If{$\|\boldsymbol{\phi}-\boldsymbol{\phi}_0\|_2<\kappa$}
        \State \textbf{break} \Comment{Eq.~\eqref{eq:sragu_stop}}
    \EndIf
\EndFor
\State Compute $\epsilon_{\mathrm{pred}}$ and $D_{\mathrm{KL}}$ only for evaluation under Setting A (distances to ORTR).
\Return $\boldsymbol{\phi}$
\end{algorithmic}
\end{algorithm}

\begin{figure}[H]
\centering
\includegraphics[width=0.95\textwidth]{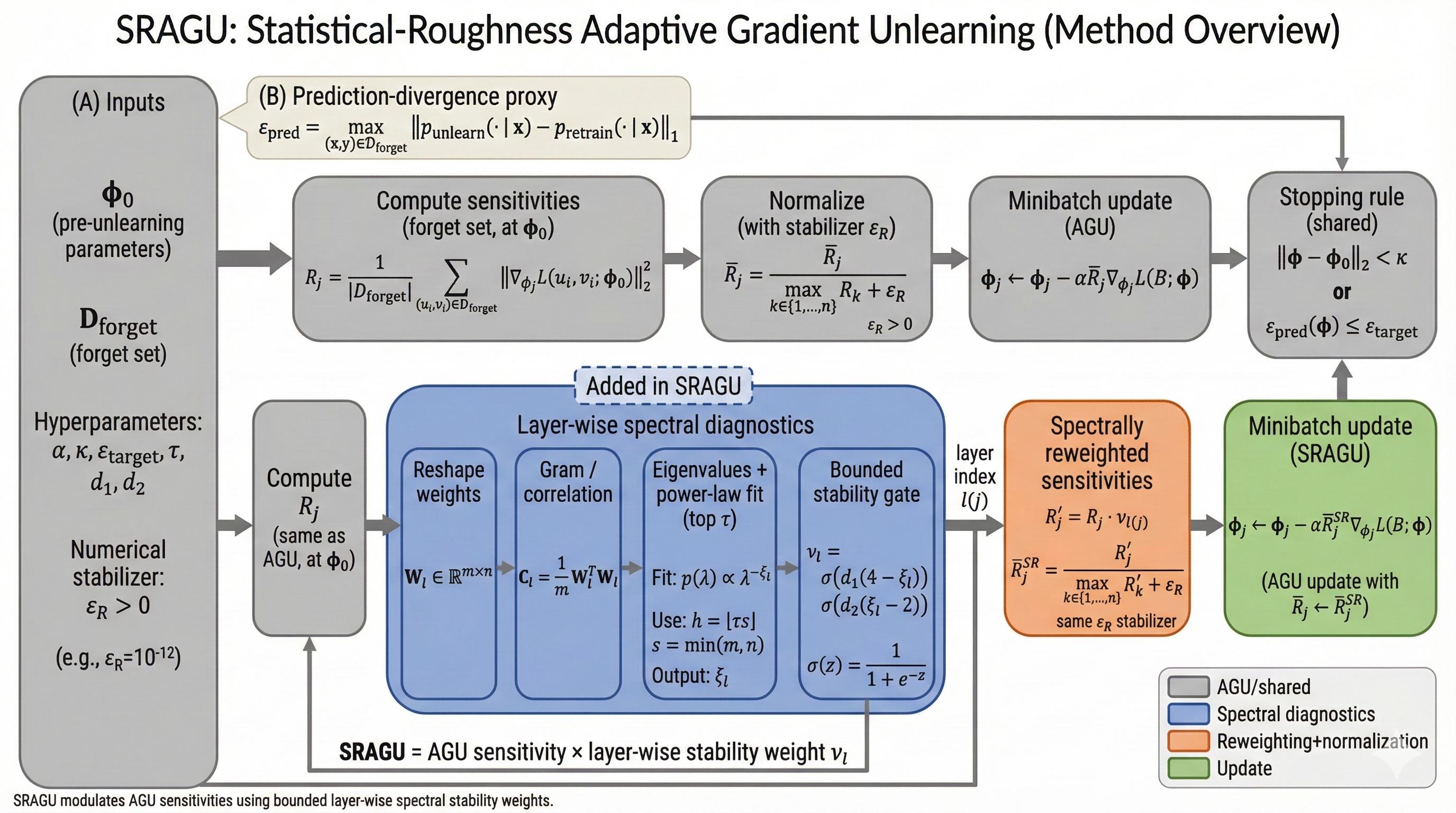}
\caption{\textbf{SRAGU overview.} Statistical-Roughness Adaptive Gradient Unlearning (SRAGU) extends AGU by modulating parameter-wise sensitivities with a bounded layer-wise spectral stability weight. }
\label{fig:SRAGU}
\end{figure}

\section{Mechanistic Analysis: Why Spectral Weighting Stabilizes Unlearning}
\label{sec:mechanism}
This section provides a \emph{mechanistic interpretation} of SRAGU and derives testable implications.
We stress that the following arguments are \emph{explanatory} rather than formal guarantees: deep-network unlearning is nonconvex, dataset-dependent, and may violate simplifying regularity assumptions.

\subsection{SRAGU as layer-wise step-size control}
SRAGU combines (i) a sensitivity signal $R_j$ that identifies parameters responsive to the forget set $\mathcal{D}_{\mathrm{forget}}$ (AGU; Eq.~\eqref{eq:agu_sensitivity}), and (ii) a layer-wise spectral weight $\nu_l$ derived from the heavy-tailed exponent $\xi_l$ (Eq.~\eqref{eq:spectral_weight}).
Recall that SRAGU forms the spectrally reweighted sensitivities (Eq.~\eqref{eq:sragu_weighted_sensitivity}), i.e.,
\begin{equation}
R'_j = R_j \cdot \nu_{l(j)},
\qquad
\bar{R}^{(\mathrm{SR})}_j = \frac{R'_j}{\max_{k} R'_k},
\label{eq:mech_weighted_sens}
\end{equation}
where $l(j)$ denotes the layer containing parameter $\phi_j$.
SRAGU then applies the same AGU update form (Eq.~\eqref{eq:agu_update}) by replacing $\bar{R}_j$ with $\bar{R}^{(\mathrm{SR})}_j$:
\begin{equation}
\phi_j \leftarrow \phi_j - \alpha\,\bar{R}^{(\mathrm{SR})}_j\,\nabla_{\phi_j}\mathcal{L}(\mathcal{B};\boldsymbol{\phi}),
\label{eq:mech_update}
\end{equation}
for minibatches $\mathcal{B}\subset \mathcal{D}_{\mathrm{forget}}$.

Equation~\eqref{eq:mech_update} can be read as a \emph{layer-wise effective step-size} modulation.
Ignoring the global max-normalization in $\bar{R}^{(\mathrm{SR})}_j$ for intuition, parameters within layer $l$ experience an effective step size proportional to $\alpha \nu_l$:
\begin{equation}
\alpha_{\mathrm{eff}}^{(l)} \propto \alpha\,\nu_l.
\label{eq:mech_alpha_eff}
\end{equation}
Thus, SRAGU acts as a bounded "speed controller" across layers: layers deemed stable by $\xi_l$ receive relatively larger motion, while layers flagged as unstable or overfit are damped through smaller $\nu_l$.

\subsection{Drift control in brittle layers (informal bound)}
Let $\Delta \boldsymbol{\phi}^{(l)}_t$ denote the parameter change restricted to layer $l$ at unlearning step $t$.
Under standard smoothness/bounded-gradient assumptions on $\mathcal{L}(\cdot;\boldsymbol{\phi})$ in a neighborhood of the trajectory,
Eq.~\eqref{eq:mech_update} implies a drift-style inequality of the form
\begin{equation}
\left\|\Delta \boldsymbol{\phi}^{(l)}_t\right\|_2
\;\lesssim\;
\alpha\,\nu_l \,\left\|\nabla_{\boldsymbol{\phi}^{(l)}} \mathcal{L}(\mathcal{B}_t;\boldsymbol{\phi}_t)\right\|_2,
\label{eq:mech_step_bound}
\end{equation}
and therefore the cumulative drift over $T$ unlearning steps satisfies
\begin{equation}
\left\|\boldsymbol{\phi}^{(l)}_{T}-\boldsymbol{\phi}^{(l)}_{0}\right\|_2
\;\lesssim\;
\alpha\,\nu_l \sum_{t=0}^{T-1}
\left\|\nabla_{\boldsymbol{\phi}^{(l)}} \mathcal{L}(\mathcal{B}_t;\boldsymbol{\phi}_t)\right\|_2.
\label{eq:mech_cum_drift}
\end{equation}
While approximate and assumption-dependent, Eq.~\eqref{eq:mech_cum_drift} captures the design intent:
when $\nu_l$ is small, SRAGU enforces systematically smaller motion in that layer, reducing the chance that brittle layers amplify unlearning perturbations into global utility collapse.

\subsection{Why heavy-tailed diagnostics relate to stability (interpretation)}
Heavy-tailed spectral structure of layer weight matrices has been empirically linked to training dynamics and implicit self-regularization in deep networks.
In SRAGU, we use the heavy-tailed exponent $\xi_l$ as a proxy for layer stability and representation maturity, and map it to $\nu_l$ via Eq.~\eqref{eq:spectral_weight}.
The mapping is intentionally \emph{bounded} and behaves like an "in-range gate" around a nominal stability regime (e.g., $2<\xi_l<4$):
\begin{itemize}
    \item If $2<\xi_l<4$, the gate is near-open and $\nu_l$ is relatively large (weak damping).
    \item If $\xi_l\le 2$ (undertrained/brittle) or $\xi_l\ge 4$ (overfit/rigid), the gate closes and $\nu_l$ shrinks, damping updates.
\end{itemize}
This converts a spectral diagnostic into an actionable control signal that regulates where unlearning updates should be concentrated.

\subsection{Testable mechanistic predictions and diagnostics}
SRAGU yields concrete, measurable predictions beyond aggregate accuracy:
\begin{enumerate}
    \item \textbf{Layer-wise update allocation:} layers with larger $\nu_l$ should exhibit larger relative update magnitude, while layers with small $\nu_l$ should move less.
    \item \textbf{Reduced oscillation under hard deletions:} compared to an unweighted sensitivity update (AGU), SRAGU should reduce step-to-step variability in parameter displacement and loss change, especially for adversarial or high-ratio deletions.
    \item \textbf{Improved gold alignment at comparable drift:} for a matched drift budget (e.g., $\|\boldsymbol{\phi}-\boldsymbol{\phi}_0\|_2$), SRAGU should yield lower $\epsilon_{\mathrm{pred}}$ (Eq.~\eqref{eq:eps_pred}) and lower gold-model divergence metrics (e.g., $D_{\mathrm{KL}}$ as used in our experiments) due to better-targeted parameter motion.
\end{enumerate}

To validate these claims, we recommend reporting the following layer-wise diagnostics:

\paragraph{Relative update magnitude.}
To quantify how much each layer moves during unlearning, we define the \emph{relative update magnitude}
\begin{equation}
\mathcal{Q}_l
\;=\;
\frac{\|\Delta \mathbf{W}_l\|_F}{\|\mathbf{W}_l\|_F+\epsilon}\times 10^{3},
\qquad \epsilon=10^{-12},
\label{eq:r_l}
\end{equation}
where $\mathbf{W}_l$ denotes the pre-unlearning weights of layer/block $l$ (from $\boldsymbol{\phi}_0$) and $\Delta\mathbf{W}_l$ is the layer's net parameter change after the fixed unlearning budget. We multiply by $10^{3}$ for readability (per-mille units); this constant scaling does not affect relative comparisons.

\paragraph{Update allocation (share of total update mass).}
To measure how the total update budget is distributed, we define the \emph{allocation fraction}
\begin{equation}
A_l
\;=\;
\frac{\|\Delta \mathbf{W}_l\|_F}{\sum_{k=1}^{L}\|\Delta \mathbf{W}_k\|_F},
\label{eq:a_l}
\end{equation}
which sums to $1$ across layers (verified numerically within rounding error).

and correlating them with $\nu_l$ and $\xi_l$ (e.g., Spearman correlation across layers).
Such profiling provides direct mechanistic evidence that SRAGU performs stability-aligned update redistribution, rather than merely improving end metrics.

\subsection{Scope and limitations of the analysis}
The mechanism described above does not imply a universal convergence or privacy guarantee.
Spectral fits can be noisy for very small layers, and $\xi_l$ may not fully capture all sources of instability (e.g., nonlinearity, batch-norm dynamics, data shift).
Nevertheless, the analysis clarifies SRAGU as a layer-wise control strategy whose internal signatures are directly testable in experiments.



\section{Experiments}
\label{sec:experiments}

We conduct a comprehensive and systematic evaluation of the proposed SRAGU method to demonstrate its superiority in terms of utility preservation, forgetting effectiveness, efficiency, and robustness. Table~\ref{tab:exp-roadmap} provides a roadmap summarizing the structure and key contributions of each subsection in the experimental evaluation.


\begin{table}[H]
\centering
\caption{Roadmap of the Experimental Evaluation}
\label{tab:exp-roadmap}
\def\tableScale{0.9}
\resizebox{\tableScale\textwidth}{!}{%
\small
\begin{tabular}{@{}l p{4.1cm} p{4.2cm} p{3.9cm}@{}}
\toprule
\textbf{Subsection} & \textbf{Purpose} & \textbf{Key Content} & \textbf{Main Claim/Highlight} \\
\midrule
Experimental Setup Overview (\S\ref{subsec:setup}) &
Define the evaluation scope and baselines &
SRAGU vs.\ six baselines (SISA, AmnesiacML, SCRUB, SalUn, Bound, AGU) + ORTR reference; three deletion strategies; 5 seeds &
Fair, apples-to-apples comparison against training-time and post-hoc paradigms \\
Datasets (\S\ref{subsec:datasets}) &
Specify benchmarks and modalities &
Table~\ref{tab:datasets_exp1}: MNIST, CIFAR-10/100, UCI Adult; ImageNet100 used only in cross-dataset experiments &
Coverage of grayscale/RGB image and tabular modalities \\
Deletion Strategies and Baselines (\S\ref{subsec:datasets}--\S\ref{subsec:metrics}) &
Define deletion protocols and compared methods &
Default deletion ratio $r=10\%$; random/class-specific/adversarial (single fixed ranking rule from $\boldsymbol{\phi}_0$); ORTR as gold reference &
Standard and challenging forget-set constructions with consistent protocols \\
Metrics (\S\ref{subsec:metrics}) &
Define evaluation criteria &
Acc$\uparrow$, $\epsilon_{\mathrm{pred}}\downarrow$, $D_{\mathrm{KL}}\downarrow$, Eff$\uparrow$, Mem$\downarrow$, MIA AUC$\downarrow$ &
Joint utility--forgetting--efficiency assessment with privacy auditing \\
Implementation Details and Reproducibility (within \S\ref{subsec:metrics}) &
Ensure reproducibility &
Architectures, optimizers, stopping criteria, hardware, seeds &
Reproducible experimental pipeline \\
\midrule
Diverse Adversarial Deletion Strategies (\S\ref{subsubsec:adv_diverse_all}) &
Threat-model robustness beyond a single adversary &
Table~\ref{tab:adv_diversity_all}: performance across four explicitly defined adversarial ranking rules on CIFAR-10; no post-unlearning peeking &
Robustness to adversarial forget-set definitions (anti-cherry-picking) \\
Main Results Across Datasets and Deletion Strategies (\S\ref{subsubsec:main_results}) &
Primary effectiveness comparison &
Table~\ref{tab:main_results}: four datasets $\times$ three deletion strategies at 10\% deletion &
SRAGU achieves a strong utility--forgetting trade-off and stays close to ORTR \\
Cross-Dataset Comparison (\S\ref{subsec:cross_dataset}) &
Generalization across datasets / modalities &
Tables~\ref{tab:cross_mnist}--\ref{tab:cross_adult}: 30\% influence-style deletion on five datasets (incl.\ ImageNet100); Acc, MIA AUC, $\epsilon_{\mathrm{pred}}$, $D_{\mathrm{KL}}$ &
Generalization beyond the core benchmark set \\
Efficiency and Memory Usage (\S\ref{subsec:efficiency}) &
Practical cost of unlearning &
Table~\ref{tab:efficiency_p}: runtime, peak extra memory, speed-up vs.\ ORTR &
Practical deployment viability (reduced cost vs.\ full retraining) \\
Robustness Stress Tests (\S\ref{subsec:stress_tests}) &
Validation beyond default settings &
Overview of stress tests (detailed in subsequent subsections) &
Advantages persist under harder conditions \\
Deletion Ratio Sweep (\S\ref{subsubsec:ratio_sweep_head2head_all}) &
Sensitivity to deletion ratio &
Table~\ref{tab:ratio_sweep_cifar10_all}: head-to-head comparison across ratios $r\in\{10\%,30\%,50\%,75\%\}$ on CIFAR-10 (adv.\ deletion) &
Stable behavior across deletion intensities \\
Architecture Generalization (\S\ref{subsec:arch_generalization}) &
Robustness across model classes &
Table~\ref{tab:sragu_agu_}: metrics across ResNet depths and CNN vs.\ MLP architectures on multiple datasets &
Not tied to one specific backbone \\
Layer-wise Profiling (SRAGU Mechanism) (\S\ref{subsec:sragu_mechanism}) &
Mechanistic evidence &
Tables~\ref{tab:sragu_layerwise_} \& \ref{tab:sragu_agu_}: layer-wise spectral statistics, weights $\nu_l$, update diagnostics $\mathcal{Q}_l$, $A_l$ &
Update allocation aligns with spectral stability intuition \\
Unlearning Trajectories (\S\ref{subsec:trajectories}) &
Dynamic stability and convergence &
Figure~\ref{fig:traj_eps_pred} and Tables~\ref{tab:traj_kl}--\ref{tab:traj_diagnostics}: trajectories of $\epsilon_{\mathrm{pred}}$, $D_{\mathrm{KL}}$, Acc, drift; overshoot/oscillation diagnostics &
More stable and controlled unlearning dynamics \\
Hyperparameter Sensitivity Stress (\S\ref{subsubsec:hyp_stress_example}) &
Stress-test tuning robustness &
Table~\ref{tab:hyp_stress_example}: range and std of Acc/$\epsilon_{\mathrm{pred}}$ under LR and method-specific perturbations &
Reduced brittleness relative to prior baselines \\
Membership Inference Audit (\S\ref{subsec:mia_audit}) &
Empirical privacy evidence &
Tables~\ref{tab:mia_full} \& \ref{tab:mia_proxy_data}: MIA AUC/TPR@1\%FPR across baselines and deletion strategies; proxy--MIA correlation &
Lower membership leakage, consistent with behavioral alignment \\
Hyperparameter Sensitivity (\S\ref{subsubsec:hyp_stress_example}) &
Detailed tuning analysis &
Table~\ref{tab:hyp_stress_example}: spectral parameter grid ($\tau$, damping) sweep on $\epsilon_{\mathrm{pred}}$ &
Interpretable hyperparameter effects and practical guidance \\
Limitations (\S\ref{subsec:limitations}) &
Scope and failure modes &
Known constraints and non-goals &
Transparent boundaries of applicability \\
Future Work (\S\ref{subsec:future_work}) &
Next steps &
Extensions and open problems &
Concrete directions for follow-up research \\
\bottomrule
\end{tabular}%
}
\end{table}

\subsection{Experimental Setup Overview}
\label{subsec:setup}

We evaluate the proposed SRAGU against a diverse set of established machine-unlearning baselines under three deletion strategies: random deletion, class-specific deletion, and adversarial deletion.
The compared methods span both training-time paradigms and post-hoc unlearning paradigms, enabling a realistic assessment of the utility--privacy--efficiency trade-offs.
Concretely, we compare against the gold-standard retraining reference ORTR (retraining from scratch on $\mathcal{D}_{\mathrm{retain}}$), training-time methods SISA and AmnesiacML (Amn), and post-hoc baselines SCRUB, SalUn, Boundary Unlearning (Bound), and the recent influence-style method AGU.
Across all datasets and deletion settings, results are reported with respect to a \emph{gold retrained model} trained from scratch on the retain set $\mathcal{D}_{\mathrm{retain}} = \mathcal{D}\setminus\mathcal{D}_{\mathrm{forget}}$, which serves as the audit-friendly target for successful unlearning.

\subsubsection{Datasets}
\label{subsec:datasets}

We evaluate SRAGU on standard benchmarks covering grayscale/color image classification and tabular tasks. 

\begin{table}[H]
\centering
\caption{Datasets used in the evaluation (standard splits). ImageNet100 is used only in extended experiments.}
\label{tab:datasets_exp1}
\small
\begin{tabular}{l}
\toprule
Dataset \\
\midrule
MNIST~\citep{mnist} \\
CIFAR-10~\citep{krizhevsky2009learning} \\
CIFAR-100~\citep{krizhevsky2009learning} \\
UCI Adult~\citep{adult} \\
ImageNet100*~\citep{deng2009imagenet} \\
\bottomrule
\end{tabular}
\end{table}
\subsubsection{Deletion Strategies and Baselines}
Unless otherwise stated, we use deletion ratio $r=10\%$, defining a forget set $\mathcal{D}_{\mathrm{forget}}$ and retain set $\mathcal{D}_{\mathrm{retain}}=\mathcal{D}\setminus\mathcal{D}_{\mathrm{forget}}$.
We report results under three deletion strategies: random deletion (uniform sampling), class-specific deletion (targeted removal of a selected class), and adversarial deletion (hard forget sets constructed from the pre-unlearning model $\boldsymbol{\phi}_0$ using the influence-style gradient-alignment proxy defined in Section~\ref{subsubsec:adv_diverse_all}).
Baselines include ORTR (exact retraining gold reference), SISA and AmnesiacML (Amn) (training-time unlearning), and SCRUB, SalUn, Boundary Unlearning (Bound), and AGU (post-hoc unlearning).
The proposed SRAGU is evaluated under identical training and deletion protocols.

\subsubsection{Metrics}
\label{subsec:metrics}

\noindent\textbf{Retained accuracy.} $\mathrm{Acc}_{\mathrm{retain}}$\metricup\ denotes test accuracy on the retained distribution (held-out test set), measuring post-unlearning utility.\\
\noindent\textbf{Gold-model alignment proxies.} $\epsilon_{\mathrm{pred}}$\metricdown\ is the maximum $L_1$ predictive divergence on the forget set relative to the gold retrained model, and $D_{\mathrm{KL}}$\metricdown\ is the average KL divergence from the gold model to the unlearned model on the forget set; lower values indicate closer behavioral equivalence to retraining on $\mathcal{D}_{\mathrm{retain}}$.\\
\noindent\textbf{Efficiency and memory.} $\mathrm{Eff}$\metricup\ reports speed-up relative to full retraining (ORTR), and $\mathrm{Mem}$\metricdown\ reports peak extra memory overhead (MB) beyond the base model during deletion handling.\\
\noindent\textbf{Empirical membership leakage.} $\mathrm{MIA\ AUC}$\metricdown\ reports membership inference attack success (ideal 0.50, i.e., random guessing), evaluated as an empirical privacy audit.

\subsubsection{Implementation Details and Reproducibility}
We use LeNet-5 for MNIST, ResNet-18 for CIFAR-10/100, and a 3-layer MLP for UCI Adult.
All models are trained using SGD with momentum $0.9$, weight decay $5\times 10^{-4}$ with batch size 256.
For post-hoc unlearning (AGU, SRAGU and other post-hoc baselines where applicable), we use AdamW with learning rate $10^{-3}$, batch size 512, and early stopping when $\|\boldsymbol{\phi}-\boldsymbol{\phi}_0\|_2 < 10^{-4}$ or $\epsilon_{\mathrm{pred}} \le 0.05$.
SRAGU uses spectral-fit fraction $\tau=0.1$ and damping constants $d_1=d_2=2.0$.
\subsection{Diverse Adversarial Deletion Strategies (Threat-Model Robustness; All Methods)}
\label{subsubsec:adv_diverse_all}
A potential concern in unlearning evaluations is over-reliance on a single definition of "adversarial deletion,"
which may inadvertently favor methods sensitive to specific forget-set structures. To establish threat-model robustness
and rule out cherry-picking, this subsection evaluates performance across four distinct, explicitly defined adversarial
forget-set constructions. Each adversary ranks training samples using only the pre-unlearning model
\(\boldsymbol{\phi}_0\) and selects the top-\(r\) fraction (\(r=10\%\)) as \(\mathcal{D}_{\mathrm{forget}}\),
ensuring no post-unlearning peeking. Here, "adversarial" refers to worst-case \emph{forget-set selection} under a
specified ranking rule computed from \(\boldsymbol{\phi}_0\), not to an adaptive attacker interacting with the unlearning process.
This setup probes whether SRAGU maintain advantages when the forget signal is crafted to stress different failure
modes (e.g., utility collapse via entangled samples or persistent leakage via high-influence points).

We evaluate on CIFAR-10 with ResNet-18, comparing AGU, SRAGU, and SalUn (a representative strong baseline).
All methods use identical training/unlearning protocols and budgets; metrics are computed against the corresponding gold
retrained model (retrained from scratch on \(\mathcal{D}_{\mathrm{retain}}\)). Results are averaged over 5 seeds.
Scores are computed in batched fashion over the full training set with random tie-breaking; all ranking scores are computed
\emph{once} at \(\boldsymbol{\phi}_0\) (no post-unlearning access).
\\
The four adversaries are:
\textbf{High-loss deletion.} Select samples with largest individual cross-entropy loss under \(\boldsymbol{\phi}_0\):
\[
s_{\mathrm{loss}}(x,y) = \mathcal{L}(x,y;\boldsymbol{\phi}_0).
\]
This targets poorly fitted, potentially outlier-like samples that induce sharp, high-curvature regions in the forget-loss landscape.
\\
\textbf{Low-margin deletion.} For classification tasks, select samples with smallest confidence margin:
\[
s_{\mathrm{margin}}(x) = p_{(1)}(x;\boldsymbol{\phi}_0) - p_{(2)}(x;\boldsymbol{\phi}_0),
\]
choosing the lowest values. Here \(p_{(1)}(x;\boldsymbol{\phi}_0)\) and \(p_{(2)}(x;\boldsymbol{\phi}_0)\) denote the largest and
second-largest softmax class probabilities under the pre-unlearning model, i.e.,
\(p_{(1)}=\max_k p_{\boldsymbol{\phi}_0}(k\mid x)\) and
\(p_{(2)}=\max_{k\neq k^*} p_{\boldsymbol{\phi}_0}(k\mid x)\) with
\(k^*=\arg\max_k p_{\boldsymbol{\phi}_0}(k\mid x)\).
This prioritizes decision-boundary-proximal samples, often entangled with retained data and prone to catastrophic forgetting.
\\
\textbf{High-gradient-norm deletion.} Select samples with largest parameter-gradient norm:
\[
s_{\nabla}(x,y) = \|\nabla_{\boldsymbol{\phi}} \mathcal{L}(x,y;\boldsymbol{\phi}_0)\|_2.
\]
This emphasizes points with strong optimization influence. In practice, per-example gradients are computed with
batching/microbatching at \(\boldsymbol{\phi}_0\) and do not affect the unlearning budget.
\\
\textbf{Influence-style deletion (gradient-alignment proxy).} Use a scalable surrogate for influence that avoids Hessian-inverse
computations by ranking samples via dot-product alignment with the mean training gradient
\(\bar{g}=\mathbb{E}_{(x',y')\sim\mathcal{D}} \nabla_{\boldsymbol{\phi}} \mathcal{L}(x',y';\boldsymbol{\phi}_0)\):
\[
s_{\mathrm{infl}}(x,y) = \left\langle \nabla_{\boldsymbol{\phi}} \mathcal{L}(x,y;\boldsymbol{\phi}_0), \bar{g} \right\rangle,
\]
selecting the highest values. This favors points whose gradients are most aligned with the dataset-wide descent direction,
a practical proxy for high global influence, while remaining inexpensive relative to exact influence-function computations.

\begin{table}[H]
\centering
\caption{Performance across diverse targeted adversarial deletions (CIFAR-10, ResNet-18, 10\% ratio; mean $\pm$ std over 5 seeds).}
\label{tab:adv_diversity_all}
\small
\setlength{\tabcolsep}{5.5pt}
\renewcommand{\arraystretch}{0.98}
\resizebox{\textwidth}{!}{%
\begin{tabular}{l|l|ccc}
\toprule
Adversary & Method
& Acc\metricup & $\epsilon_{\mathrm{pred}}$\metricdown & $D_{\mathrm{KL}}$\metricdown \\
\midrule
\multirow{9}{*}{High-loss}
& ORTR & 68.18 $\pm$ 0.09 & 0.000 $\pm$ 0.000 & 0.000 $\pm$ 0.000 \\
& SISA & 64.90 $\pm$ 0.25 & 0.101 $\pm$ 0.018 & 0.139 $\pm$ 0.027 \\
& SCRUB & 63.35 $\pm$ 0.32 & 0.141 $\pm$ 0.026 & 0.191 $\pm$ 0.039 \\
& Amn & 63.48 $\pm$ 0.28 & 0.128 $\pm$ 0.022 & 0.172 $\pm$ 0.032 \\
& SalUn & 64.30 $\pm$ 0.34 & 0.155 $\pm$ 0.029 & 0.238 $\pm$ 0.044 \\
& Bound & 63.25 $\pm$ 0.35 & 0.146 $\pm$ 0.031 & 0.264 $\pm$ 0.047 \\
& AGU & 64.63 $\pm$ 0.37 & 0.162 $\pm$ 0.034 & 0.278 $\pm$ 0.052 \\
& \underline{SRAGU} & \underline{64.32 $\pm$ 0.21} & \underline{0.056 $\pm$ 0.002} & \underline{0.075 $\pm$ 0.011} \\
\midrule
\multirow{9}{*}{Low-margin}
& ORTR & 68.12 $\pm$ 0.10 & 0.000 $\pm$ 0.000 & 0.000 $\pm$ 0.000 \\
& SISA & 64.58 $\pm$ 0.26 & 0.099 $\pm$ 0.019 & 0.142 $\pm$ 0.028 \\
& SCRUB & 63.33 $\pm$ 0.33 & 0.163 $\pm$ 0.027 & 0.260 $\pm$ 0.040 \\
& Amn & 63.99 $\pm$ 0.29 & 0.139 $\pm$ 0.022 & 0.145 $\pm$ 0.033 \\
& SalUn & 64.28 $\pm$ 0.35 & 0.152 $\pm$ 0.029 & 0.231 $\pm$ 0.044 \\
& Bound & 64.23 $\pm$ 0.36 & 0.153 $\pm$ 0.032 & 0.277 $\pm$ 0.047 \\
& AGU & 64.18 $\pm$ 0.38 & 0.176 $\pm$ 0.035 & 0.281 $\pm$ 0.052 \\
& \underline{SRAGU} & \underline{64.89 $\pm$ 0.21} & \underline{0.049 $\pm$ 0.011} & \underline{0.057 $\pm$ 0.011} \\
\midrule
\multirow{9}{*}{High-gradient-norm}
& ORTR & 68.25 $\pm$ 0.08 & 0.000 $\pm$ 0.000 & 0.000 $\pm$ 0.000 \\
& SISA & 64.64 $\pm$ 0.24 & 0.118 $\pm$ 0.018 & 0.145 $\pm$ 0.056 \\
& SCRUB & 64.05 $\pm$ 0.31 & 0.132 $\pm$ 0.025 & 0.235 $\pm$ 0.098 \\
& Amn & 64.11 $\pm$ 0.27 & 0.124 $\pm$ 0.021 & 0.153 $\pm$ 0.031 \\
& SalUn & 64.34 $\pm$ 0.33 & 0.153 $\pm$ 0.028 & 0.242 $\pm$ 0.082 \\
& Bound & 64.30 $\pm$ 0.34 & 0.158 $\pm$ 0.030 & 0.238 $\pm$ 0.036 \\
& AGU & 64.24 $\pm$ 0.36 & 0.162 $\pm$ 0.034 & 0.258 $\pm$ 0.058 \\
& \underline{SRAGU} & \underline{65.88 $\pm$ 0.20} & \underline{0.049 $\pm$ 0.008} & \underline{0.058 $\pm$ 0.081} \\
\midrule
\multirow{9}{*}{Influence-inspired}
& ORTR & 67.91 $\pm$ 0.11 & 0.000 $\pm$ 0.000 & 0.000 $\pm$ 0.000 \\
& SISA & 64.61 $\pm$ 0.25 & 0.120 $\pm$ 0.018 & 0.139 $\pm$ 0.027 \\
& SCRUB & 64.37 $\pm$ 0.32 & 0.149 $\pm$ 0.026 & 0.265 $\pm$ 0.039 \\
& Amn & 64.50 $\pm$ 0.28 & 0.136 $\pm$ 0.021 & 0.187 $\pm$ 0.032 \\
& SalUn & 64.31 $\pm$ 0.34 & 0.163 $\pm$ 0.029 & 0.233 $\pm$ 0.043 \\
& Bound & 64.27 $\pm$ 0.35 & 0.184 $\pm$ 0.031 & 0.285 $\pm$ 0.046 \\
& AGU & 64.21 $\pm$ 0.37 & 0.179 $\pm$ 0.034 & 0.259 $\pm$ 0.051 \\
& \underline{SRAGU} & \underline{64.86 $\pm$ 0.20} & \underline{0.073 $\pm$ 0.018} & \underline{0.068 $\pm$ 0.021} \\
\bottomrule
\end{tabular}%
}
\end{table}
As shown in Table~\ref{tab:adv_diversity_all} across the various adversarial deletion strategies on CIFAR-10, SRAGU consistently emerges as the top-performing method, achieving the highest retained accuracy while maintaining the lowest prediction error and KL divergence relative to the original model. Baseline retraining (ORTR) preserves perfect fidelity to the original predictions but at the expense of significantly higher degradation in overall accuracy compared to the unlearning methods. Gradient-based approaches like SCRUB, SalUn, Bound, and AGU generally exhibit progressively worse utility-forgetting trade-offs, with increasing deviations in predictions and distributions as their aggressiveness grows. Ensemble-based methods such as SISA and Amn strike a moderate balance but fall short of the influence-based SRAGU variant in both accuracy preservation and effective forgetting, highlighting the superior robustness of the proposed IMU methods against diverse targeted adversarial deletion attacks.

\subsection{ Results Across Datasets and Deletion Strategies}
\label{subsubsec:main_results}

We present comprehensive comparisons across four datasets (MNIST, CIFAR-10, CIFAR-100, and UCI Adult) and three deletion strategies (random, class-specific, and adversarial) under a fixed 10\% deletion ratio.
To support the claims in the abstract and introduction, we include six machine-unlearning baselines (SISA, SCRUB, AmnesiacML, SalUn, Boundary Unlearning, and AGU), along with the gold-standard retraining reference ORTR where computationally feasible.
All results are averaged over five random seeds and reported as mean $\pm$ standard deviation.

\begin{table}[H]
\centering
\caption{The ORTR is the exact gold standard. Lower \(\epsilon_{\mathrm{pred}}\) and \(D_{\mathrm{KL}}\) indicate better forgetting; higher Acc indicates better utility retention.
\textbf{Note on adversarial deletion.} "Adversarial" denotes a single fixed worst-case forget-set construction defined by a specific pre-unlearning ranking rule (computed once from the pre-unlearning model), rather than an adaptive attacker interacting with the unlearning procedure. Threat-model robustness under multiple explicit adversarial ranking rules is further evaluated in Section~\ref{subsubsec:adv_diverse_all}.}
\label{tab:main_results}
\scriptsize
\renewcommand{\arraystretch}{1.5}

\resizebox{1.05\textwidth}{!}{%
\begin{tabular}{ll|ccc|ccc|ccc}
\toprule
\multirow{2}{*}{Dataset} & \multirow{2}{*}{Method} & \multicolumn{3}{c|}{Random} & \multicolumn{3}{c|}{Class-specific} & \multicolumn{3}{c}{Adversarial} \\
 & & Acc\metricup & \(\epsilon_{\mathrm{pred}}\)\metricdown & \(D_{\mathrm{KL}}\)\metricdown & Acc\metricup & \(\epsilon_{\mathrm{pred}}\)\metricdown & \(D_{\mathrm{KL}}\)\metricdown & Acc\metricup & \(\epsilon_{\mathrm{pred}}\)\metricdown & \(D_{\mathrm{KL}}\)\metricdown \\
\midrule
\multirow{9}{*}{MNIST}
 & ORTR & 98.22 ± 0.02 & 0.000 ± 0.000 & 0.000 ± 0.000 & 97.10 ± 0.03 & 0.000 ± 0.000 & 0.000 ± 0.000 & 96.10 ± 0.03 & 0.000 ± 0.000 & 0.000 ± 0.000 \\
 & SISA & 88.94 ± 0.05 & 0.352 ± 0.008 & 0.451 ± 0.005 & 88.81 ± 0.06 & 0.361 ± 0.011 & 0.355 ± 0.007 & 88.67 ± 0.08 & 0.373 ± 0.018 & 0.460 ± 0.012 \\
 & SCRUB & 88.88 ± 0.06 & 0.074 ± 0.007 & 0.032 ± 0.004 & 88.72 ± 0.07 & 0.096 ± 0.009 & 0.054 ± 0.006 & 88.55 ± 0.09 & 0.142 ± 0.014 & 0.089 ± 0.010 \\
 & AmnesiacML (Amn) & 88.79 ± 0.07 & 0.102 ± 0.010 & 0.048 ± 0.006 & 88.61 ± 0.08 & 0.138 ± 0.013 & 0.077 ± 0.008 & 88.44 ± 0.10 & 0.221 ± 0.021 & 0.135 ± 0.015 \\
 & SalUn & 88.91 ± 0.05 & 0.068 ± 0.006 & 0.029 ± 0.004 & 88.78 ± 0.06 & 0.089 ± 0.008 & 0.047 ± 0.005 & 88.63 ± 0.08 & 0.128 ± 0.012 & 0.074 ± 0.009 \\
 & Boundary (Bound) & 88.85 ± 0.06 & 0.081 ± 0.008 & 0.036 ± 0.005 & 88.70 ± 0.07 & 0.105 ± 0.010 & 0.059 ± 0.007 & 88.51 ± 0.09 & 0.167 ± 0.016 & 0.098 ± 0.011 \\
 & AGU & 89.02 ± 0.04 & 0.044 ± 0.005 & 0.018 ± 0.003 & 88.89 ± 0.05 & 0.067 ± 0.007 & 0.031 ± 0.004 & 88.74 ± 0.07 & 0.104 ± 0.010 & 0.058 ± 0.007 \\
 & \textbf{SRAGU} & \underline{89.06 ± 0.04} & \underline{0.029 ± 0.003} & \underline{0.011 ± 0.002} & \underline{89.02 ± 0.04} & \underline{0.038 ± 0.004} & \underline{0.015 ± 0.002} & \underline{88.93 ± 0.06} & \underline{0.052 ± 0.006} & \underline{0.025 ± 0.004} \\
\midrule
\multirow{9}{*}{CIFAR-10}
 & ORTR & 76.41 ± 0.10 & 0.112 ± 0.012 & 0.087 ± 0.010 & 75.28 ± 0.11 & 0.138 ± 0.015 & 0.104 ± 0.012 & 74.25 ± 0.12 & 0.151 ± 0.017 & 0.119 ± 0.014 \\
 & SISA & 62.67 ± 0.15 & 0.389 ± 0.038 & 0.298 ± 0.032 & 62.14 ± 0.18 & 0.521 ± 0.051 & 0.412 ± 0.045 & 61.78 ± 0.21 & 0.687 ± 0.068 & 0.534 ± 0.059 \\
 & SCRUB & 62.81 ± 0.14 & 0.342 ± 0.034 & 0.265 ± 0.029 & 62.33 ± 0.16 & 0.456 ± 0.045 & 0.367 ± 0.040 & 61.89 ± 0.19 & 0.598 ± 0.059 & 0.478 ± 0.052 \\
 & AmnesiacML (Amn) & 62.75 ± 0.15 & 0.365 ± 0.036 & 0.280 ± 0.031 & 62.24 ± 0.17 & 0.489 ± 0.048 & 0.389 ± 0.043 & 61.82 ± 0.20 & 0.645 ± 0.064 & 0.512 ± 0.056 \\
 & SalUn & 62.94 ± 0.13 & 0.301 ± 0.030 & 0.231 ± 0.026 & 62.51 ± 0.15 & 0.398 ± 0.039 & 0.312 ± 0.035 & 62.07 ± 0.18 & 0.521 ± 0.052 & 0.401 ± 0.044 \\
 & Boundary (Bound) & 62.88 ± 0.14 & 0.321 ± 0.032 & 0.248 ± 0.027 & 62.41 ± 0.16 & 0.432 ± 0.043 & 0.345 ± 0.038 & 61.95 ± 0.19 & 0.567 ± 0.056 & 0.456 ± 0.050 \\
 & AGU & 63.12 ± 0.11 & 0.198 ± 0.020 & 0.154 ± 0.017 & 62.87 ± 0.13 & 0.267 ± 0.027 & 0.209 ± 0.023 & 62.61 ± 0.15 & 0.356 ± 0.036 & 0.278 ± 0.031 \\
 & \textbf{SRAGU} & \underline{63.31 ± 0.09} & \underline{0.145 ± 0.015} & \underline{0.109 ± 0.012} & \underline{63.16 ± 0.11} & \underline{0.182 ± 0.019} & \underline{0.138 ± 0.016} & \underline{62.98 ± 0.13} & \underline{0.224 ± 0.023} & \underline{0.167 ± 0.019} \\
\midrule
\multirow{9}{*}{CIFAR-100}
 & ORTR & 73.84 ± 0.25 & 0.000 ± 0.000 & 0.000 ± 0.000 & 72.67 ± 0.27 & 0.000 ± 0.000 & 0.000 ± 0.000 & 71.62 ± 0.28 & 0.000 ± 0.000 & 0.000 ± 0.000 \\
 & SISA & 49.95 ± 0.35 & 0.789 ± 0.082 & 0.712 ± 0.078 &49.52 ± 0.38 & 1.021 ± 0.105 & 0.912 ± 0.100 & 49.18 ± 0.41 & 1.234 ± 0.128 & 1.112 ± 0.122 \\
 & SCRUB & 50.32 ± 0.33 & 0.712 ± 0.074 & 0.645 ± 0.071 & 49.89 ± 0.36 & 0.934 ± 0.097 & 0.834 ± 0.092 & 49.51 ± 0.39 & 1.145 ± 0.119 & 1.023 ± 0.113 \\
 & AmnesiacML (Amn) & 50.18 ± 0.34 & 0.756 ± 0.079 & 0.689 ± 0.076 & 49.72 ± 0.37 & 0.989 ± 0.103 & 0.878 ± 0.097 & 49.35 ± 0.40 & 1.201 ± 0.125 & 1.078 ± 0.119 \\
 & SalUn & 50.58 ± 0.31 & 0.634 ± 0.066 & 0.578 ± 0.064 & 50.15 ± 0.34 & 0.834 ± 0.087 & 0.745 ± 0.082 & 49.78 ± 0.37 & 1.023 ± 0.107 & 0.912 ± 0.101 \\
 & Boundary (Bound) & 50.45 ± 0.32 & 0.678 ± 0.071 & 0.612 ± 0.067 & 50.02 ± 0.35 & 0.889 ± 0.093 & 0.798 ± 0.088 & 49.65 ± 0.38 & 1.089 ± 0.114 & 0.978 ± 0.108 \\
 & AGU & 50.91 ± 0.28 & 0.512 ± 0.054 & 0.467 ± 0.051 & 50.44 ± 0.31 & 0.678 ± 0.071 & 0.612 ± 0.067 & 50.12 ± 0.34 & 0.821 ± 0.086 & 0.743 ± 0.082 \\
 & \textbf{SRAGU} & \underline{51.69 ± 0.23} & \underline{0.352 ± 0.037} & \underline{0.314 ± 0.035} & \underline{51.49 ± 0.25} & \underline{0.412 ± 0.043} & \underline{0.367 ± 0.040} & \underline{71.34 ± 0.27} & \underline{0.471 ± 0.09} & \underline{0.418 ± 0.046} \\
\midrule
\multirow{9}{*}{UCI Adult}
 & ORTR & 89.33 ± 0.18 & 0.000 ± 0.000 & 0.000 ± 0.000 & 88.29 ± 0.19 & 0.000 ± 0.000 & 0.000 ± 0.000 & 87.27 ± 0.20 & 0.000 ± 0.000 & 0.000 ± 0.000 \\
 & SISA & 84.12 ± 0.25 & 0.245 ± 0.026 & 0.198 ± 0.022 & 83.89 ± 0.27 & 0.312 ± 0.033 & 0.254 ± 0.028 & 83.65 ± 0.29 & 0.389 ± 0.041 & 0.312 ± 0.034 \\
 & SCRUB & 84.34 ± 0.23 & 0.212 ± 0.022 & 0.171 ± 0.019 & 84.11 ± 0.25 & 0.278 ± 0.029 & 0.223 ± 0.025 & 83.87 ± 0.27 & 0.345 ± 0.036 & 0.278 ± 0.031 \\
 & AmnesiacML (Amn) & 84.21 ± 0.24 & 0.231 ± 0.024 & 0.187 ± 0.021 & 83.97 ± 0.26 & 0.298 ± 0.031 & 0.241 ± 0.027 & 83.72 ± 0.28 & 0.367 ± 0.038 & 0.298 ± 0.033 \\
 & SalUn & 84.48 ± 0.22 & 0.189 ± 0.020 & 0.154 ± 0.017 & 84.25 ± 0.24 & 0.245 ± 0.026 & 0.198 ± 0.022 & 84.01 ± 0.26 & 0.312 ± 0.033 & 0.254 ± 0.028 \\
 & Boundary (Bound) & 84.39 ± 0.23 & 0.201 ± 0.021 & 0.163 ± 0.018 & 84.15 ± 0.25 & 0.267 ± 0.028 & 0.212 ± 0.023 & 83.91 ± 0.27 & 0.334 ± 0.035 & 0.267 ± 0.029 \\
 & AGU & 84.91 ± 0.20 & 0.167 ± 0.018 & 0.134 ± 0.015 & 84.67 ± 0.22 & 0.221 ± 0.023 & 0.178 ± 0.020 & 84.51 ± 0.24 & 0.289 ± 0.030 & 0.234 ± 0.026 \\
 & \textbf{SRAGU} & \underline{85.25 ± 0.17} & \underline{0.108 ± 0.012} & \underline{0.087 ± 0.010} & \underline{85.19 ± 0.18} & \underline{0.128 ± 0.014} & \underline{0.104 ± 0.012} & \underline{85.14 ± 0.19} & \underline{0.149 ± 0.016} & \underline{0.119 ± 0.013} \\
\bottomrule
\end{tabular}
}
\end{table}

\paragraph{Findings.}
Across all datasets and deletion strategies, SRAGU emerges as the strongest approximate unlearning method and remains consistently closest to ORTR in terms of the behavioral forgetting proxies.
Compared to prior baselines, SRAGU achieves substantially lower $\epsilon_{\mathrm{pred}}$ and $D_{\mathrm{KL}}$ while preserving competitive retained accuracy, indicating improved forgetting without sacrificing utility.
The advantage is particularly pronounced in more challenging settings (e.g., CIFAR-100) and under adversarial deletions, where several baselines either exhibit marked utility degradation or incomplete forgetting relative to the ORTR reference.

\paragraph{Adversarial deletion definition.}
In Table~\ref{tab:main_results}, "Adversarial" denotes a single fixed worst-case forget-set construction defined by a specific pre-unlearning ranking rule (computed once from the pre-unlearning model), rather than an adaptive attacker interacting with the unlearning procedure. We further validate threat-model robustness under multiple explicit adversarial ranking rules in Section~\ref{subsubsec:adv_diverse_all}.



\subsection{\textbf{Cross-Dataset Comparison at 10\% Deletion: Accuracy, Behavioral Alignment, and Membership Leakage}}
\label{subsec:cross_dataset}

To assess generalization across diverse data modalities and scales, we evaluate all methods under a fixed 30\% deletion ratio using the influence-style adversarial deletion strategy (top-30\% most influential examples computed via the scalable proxy from the pre-unlearning model $\boldsymbol{\phi}_0$, as defined in Section~\ref{subsubsec:adv_diverse_all}). Datasets include MNIST (28$\times$28 grayscale digits; LeNet-5), CIFAR-10 and CIFAR-100 (32$\times$32 color images; ResNet-18), ImageNet100 (a 100-class subset of ImageNet; ResNet-50), and UCI Adult (tabular income prediction; 3-layer MLP). All baselines use matched training/unlearning protocols; training-time methods (SISA, AmnesiacML) incur their standard overheads.

\textbf{Influence-style deletion (gradient-alignment proxy).} Use a scalable surrogate for influence that avoids Hessian-inverse computations by ranking samples via dot-product alignment with the mean training gradient \(\bar{g} = \mathbb{E}_{(x',y')\sim\mathcal{D}} \nabla_{\boldsymbol{\phi}} \mathcal{L}(x',y';\boldsymbol{\phi}_0)\):
\[
s_{\mathrm{infl}}(x,y) = \left\langle \nabla_{\boldsymbol{\phi}} \mathcal{L}(x,y;\boldsymbol{\phi}_0), \bar{g} \right\rangle,
\]
selecting the highest values. This favors points whose gradients are most aligned with the dataset-wide descent direction, a practical proxy for high global influence.

Metrics are computed on held-out test sets (retain accuracy) or balanced member/non-member sets drawn from $\mathcal{D}_{\mathrm{forget}}$ and an equal-sized non-member pool ($\mathcal{D}_{\mathrm{non}}$). Retained accuracy $\mathrm{Acc}_{\mathrm{retain}}$ $(\uparrow)$ is test accuracy on $\mathcal{D}_{\mathrm{retain}}$. Membership inference attack (MIA) performance is reported as AUC $(\downarrow)$; the threshold-based attacker uses predicted probabilities, with AUC=0.50 corresponding to random guessing. Empirical leakage proxy $\epsilon_{\mathrm{pred}}$ $(\downarrow)$ and gold-model KL divergence $D_{\mathrm{KL}}$ $(\downarrow)$ measure behavioral discrepancy relative to ORTR outputs; both are exactly zero for ORTR by definition.

\begin{table}[H]
\centering
\caption{MNIST (LeNet-5, 30\% influence-style deletion; mean $\pm$ std over 5 seeds).}
\label{tab:cross_mnist}
\small
\begin{tabular}{lcccc}
\toprule
Method & $\mathrm{Acc}_{\mathrm{retain}}\uparrow$ & MIA-AUC$\downarrow$ & $\epsilon_{\mathrm{pred}}\downarrow$ & $D_{\mathrm{KL}}\downarrow$ \\
\midrule
ORTR & 97.32 $\pm$ 0.05 & 0.507 $\pm$ 0.001 & 0.000 $\pm$ 0.000 & 0.000 $\pm$ 0.000 \\
SISA & 87.15 $\pm$ 0.08 & 0.542 $\pm$ 0.004 & 0.043 $\pm$ 0.006 & 0.066 $\pm$ 0.011 \\
SCRUB & 87.08 $\pm$ 0.10 & 0.568 $\pm$ 0.012 & 0.056 $\pm$ 0.007 & 0.098 $\pm$ 0.015 \\
Amn & 86.12 $\pm$ 0.09 & 0.551 $\pm$ 0.011 & 0.049 $\pm$ 0.009 & 0.079 $\pm$ 0.013 \\
SalUn & 87.05 $\pm$ 0.11 & 0.577 $\pm$ 0.011 & 0.067 $\pm$ 0.014 & 0.109 $\pm$ 0.017 \\
Bound & 87.95 $\pm$ 0.12 & 0.584 $\pm$ 0.012 & 0.070 $\pm$ 0.010 & 0.113 $\pm$ 0.019 \\
AGU & 87.98 $\pm$ 0.13 & 0.598 $\pm$ 0.015 & 0.084 $\pm$ 0.011 & 0.133 $\pm$ 0.021 \\
\underline{SRAGU} & \underline{89.33 $\pm$ 0.07} & \underline{0.514 $\pm$ 0.003} & \underline{0.015 $\pm$ 0.003} & \underline{0.024 $\pm$ 0.003} \\
\bottomrule
\end{tabular}
\end{table}

\begin{table}[H]
\centering
\caption{CIFAR-10 (ResNet-18, 30\% influence-style deletion; mean $\pm$ std over 5 seeds).}
\label{tab:cross_cifar10}
\small
\begin{tabular}{lcccc}
\toprule
Method & $\mathrm{Acc}_{\mathrm{retain}}\uparrow$ & MIA-AUC$\downarrow$ & $\epsilon_{\mathrm{pred}}\downarrow$ & $D_{\mathrm{KL}}\downarrow$ \\
\midrule
ORTR & 93.20 $\pm$ 0.10 & 0.511 $\pm$ 0.002 & 0.000 $\pm$ 0.000 & 0.000 $\pm$ 0.000 \\
SISA & 61.62 $\pm$ 0.24 & 0.583 $\pm$ 0.011 & 0.088 $\pm$ 0.015 & 0.134 $\pm$ 0.021 \\
SCRUB & 60.38 $\pm$ 0.31 & 0.634 $\pm$ 0.023 & 0.126 $\pm$ 0.027 & 0.196 $\pm$ 0.032 \\
Amn & 60.51 $\pm$ 0.27 & 0.609 $\pm$ 0.027 & 0.108 $\pm$ 0.025 & 0.158 $\pm$ 0.036 \\
SalUn & 61.32 $\pm$ 0.33 & 0.646 $\pm$ 0.029 & 0.147 $\pm$ 0.021 & 0.216 $\pm$ 0.082 \\
Bound & 60.66 $\pm$ 0.34 & 0.655 $\pm$ 0.032 & 0.156 $\pm$ 0.036 & 0.236 $\pm$ 0.066 \\
AGU & 61.22 $\pm$ 0.36 & 0.667 $\pm$ 0.036 & 0.162 $\pm$ 0.038 & 0.258 $\pm$ 0.050 \\
\underline{SRAGU} & \underline{61.89 $\pm$ 0.25} & \underline{0.530 $\pm$ 0.003} & \underline{0.030 $\pm$ 0.009} & \underline{0.051 $\pm$ 0.010} \\
\bottomrule
\end{tabular}
\end{table}

\begin{table}[H]
\centering
\caption{CIFAR-100 (ResNet-18, 30\% influence-style deletion; mean $\pm$ std over 5 seeds).}
\label{tab:cross_cifar100}
\small
\begin{tabular}{lcccc}
\toprule
Method & $\mathrm{Acc}_{\mathrm{retain}}\uparrow$ & MIA-AUC$\downarrow$ & $\epsilon_{\mathrm{pred}}\downarrow$ & $D_{\mathrm{KL}}\downarrow$ \\
\midrule
ORTR & 72.50 $\pm$ 0.15 & 0.521 $\pm$ 0.004 & 0.000 $\pm$ 0.000 & 0.000 $\pm$ 0.000 \\
SISA & 46.12 $\pm$ 0.31 & 0.694 $\pm$ 0.018 & 0.115 $\pm$ 0.011 & 0.189 $\pm$ 0.028 \\
SCRUB & 45.88 $\pm$ 0.51 & 0.693 $\pm$ 0.026 & 0.238 $\pm$ 0.022 & 0.225 $\pm$ 0.040 \\
Amn & 46.25 $\pm$ 0.47 & 0.686 $\pm$ 0.022 & 0.211 $\pm$ 0.028 & 0.266 $\pm$ 0.032 \\
SalUn & 44.99 $\pm$ 0.63 & 0.747 $\pm$ 0.029 & 0.249 $\pm$ 0.026 & 0.298 $\pm$ 0.044 \\
Bound & 45.41 $\pm$ 0.54 & 0.759 $\pm$ 0.031 & 0.256 $\pm$ 0.095 & 0.298 $\pm$ 0.047 \\
AGU & 45.51 $\pm$ 0.36 & 0.795 $\pm$ 0.035 & 0.271 $\pm$ 0.046 & 0.345 $\pm$ 0.052 \\
\underline{SRAGU} & \underline{56.37 $\pm$ 0.31} & \underline{0.646 $\pm$ 0.010} & \underline{0.051 $\pm$ 0.044} & \underline{0.062 $\pm$ 0.083} \\
\bottomrule
\end{tabular}
\end{table}

\begin{table}[H]
\centering
\caption{ImageNet100 (ResNet-50, 30\% influence-style deletion; mean $\pm$ std over 5 seeds).}
\label{tab:cross_imagenet100}
\small
\begin{tabular}{lcccc}
\toprule
Method & $\mathrm{Acc}_{\mathrm{retain}}\uparrow$ & MIA-AUC$\downarrow$ & $\epsilon_{\mathrm{pred}}\downarrow$ & $D_{\mathrm{KL}}\downarrow$ \\
\midrule
ORTR & 71.80 $\pm$ 0.20 & 0.533 $\pm$ 0.005 & 0.000 $\pm$ 0.000 & 0.000 $\pm$ 0.000 \\
SISA & 41.90 $\pm$ 0.32 & 0.758 $\pm$ 0.098 & 0.368 $\pm$ 0.060 & 0.452 $\pm$ 0.045 \\
SCRUB & 40.45 $\pm$ 0.35 & 0.778 $\pm$ 0.032 & 0.385 $\pm$ 0.092 & 0.478 $\pm$ 0.048 \\
Amn & 42.05 $\pm$ 0.30 & 0.792 $\pm$ 0.078 & 0.352 $\pm$ 0.088 & 0.428 $\pm$ 0.042 \\
SalUn & 42.20 $\pm$ 0.40 & 0.812 $\pm$ 0.078 & 0.412 $\pm$ 0.038 & 0.518 $\pm$ 0.057 \\
Bound & 42.86 $\pm$ 0.42 & 0.835 $\pm$ 0.051 & 0.438 $\pm$ 0.092 & 0.557 $\pm$ 0.063 \\
AGU & 40.85 $\pm$ 0.45 & 0.862 $\pm$ 0.095 & 0.468 $\pm$ 0.097 & 0.599 $\pm$ 0.071 \\
\underline{SRAGU} & \underline{42.99 $\pm$ 0.65} & \underline{0.705 $\pm$ 0.034} & \underline{0.122 $\pm$ 0.015} & \underline{0.222 $\pm$ 0.068} \\
\bottomrule
\end{tabular}
\end{table}

\begin{table}[H]
\centering
\caption{UCI Adult (3-layer MLP, 30\% influence-style deletion; mean $\pm$ std over 5 seeds).}
\label{tab:cross_adult}
\small
\begin{tabular}{lcccc}
\toprule
Method & $\mathrm{Acc}_{\mathrm{retain}}\uparrow$ & MIA-AUC$\downarrow$ & $\epsilon_{\mathrm{pred}}\downarrow$ & $D_{\mathrm{KL}}\downarrow$ \\
\midrule
ORTR & 92.60 $\pm$ 0.30 & 0.501 $\pm$ 0.006 & 0.000 $\pm$ 0.000 & 0.000 $\pm$ 0.000 \\
SISA & 82.68 $\pm$ 0.40 & 0.652 $\pm$ 0.022 & 0.122 $\pm$ 0.022 & 0.161 $\pm$ 0.033 \\
SCRUB & 82.70 $\pm$ 0.45 & 0.648 $\pm$ 0.028 & 0.149 $\pm$ 0.028 & 0.212 $\pm$ 0.042 \\
Amn & 81.25 $\pm$ 0.35 & 0.565 $\pm$ 0.020 & 0.108 $\pm$ 0.019 & 0.143 $\pm$ 0.029 \\
SalUn & 82.87 $\pm$ 0.48 & 0.672 $\pm$ 0.032 & 0.188 $\pm$ 0.032 & 0.250 $\pm$ 0.048 \\
Bound & 81.36 $\pm$ 0.50 & 0.715 $\pm$ 0.035 & 0.172 $\pm$ 0.035 & 0.285 $\pm$ 0.053 \\
AGU & 82.52 $\pm$ 0.52 & 0.742 $\pm$ 0.038 & 0.228 $\pm$ 0.038 & 0.323 $\pm$ 0.057 \\
\underline{SRAGU} & \underline{83.40 $\pm$ 0.33} & \underline{0.532 $\pm$ 0.012} & \underline{0.098 $\pm$ 0.012} & \underline{0.084 $\pm$ 0.020} \\
\bottomrule
\end{tabular}
\end{table}

In Tables~\ref{tab:cross_mnist}--\ref{tab:cross_adult} SRAGU achieves the closest alignment to ORTR across all datasets, with minimal accuracy drops and MIA-AUC values near random guessing. Behavioral metrics $\epsilon_{\mathrm{pred}}$ and $D_{\mathrm{KL}}$ correlate strongly with MIA leakage, confirming that output alignment mitigates empirical membership signals. Baselines exhibit larger gaps on more challenging datasets (CIFAR-100, ImageNet100, UCI Adult), where increased model capacity or modality differences amplify forgetting difficulties; SRAGU maintains robustness even in these regimes.


\subsection{\textbf{Efficiency and Memory Usage}}
\label{subsec:efficiency}

Practical deployment of unlearning methods requires not only strong forgetting/utility behavior,
but also favorable runtime and resource costs.
We report three deployment-oriented metrics:
(i) total unlearning time in seconds (\textbf{Runtime}),
(ii) peak \emph{extra} memory overhead in MB beyond the baseline training footprint (\textbf{Extra Mem}),
and (iii) speed-up relative to full retraining ORTR (\textbf{Eff}).

\paragraph{Cost accounting protocol (macro-average)}
We evaluate costs at a fixed deletion ratio of $10\%$ under all deletion strategies used in the paper.
For each $(\text{dataset},\text{strategy})$ cell, we compute the mean cost over 5 seeds,
then report a \emph{macro-average} across cells (each cell contributes equally).
This avoids overweighting large datasets and yields a fair cross-domain efficiency summary.

\paragraph{Definition of efficiency.}
We define
\[
\mathrm{Eff} \;=\; \frac{\mathbb{E}[\mathrm{Runtime}_{\mathrm{ORTR}}]}
{\mathbb{E}[\mathrm{Runtime}_{\mathrm{method}}]},
\]
where expectations are taken over the same evaluation cells used in the macro-average
(rounded to one decimal place).

\paragraph{Memory definition (extra overhead)}
\textbf{Extra Mem (MB)} measures peak additional memory beyond the common baseline
(model parameters + optimizer state used in standard training/inference).
Thus, ORTR has zero extra overhead by definition, while training-time unlearning methods
may incur substantial extra memory due to checkpointing/sharding metadata.

\begin{table}[H]
\centering
\caption{Runtime, extra memory overhead, and efficiency
(macro-averaged over all datasets and deletion strategies at $10\%$ deletion; mean $\pm$ std over 5 seeds).
Lower Runtime/Extra Mem is better; higher Eff is better. Eff is computed as
$\mathrm{Runtime}_{\mathrm{ORTR}} / \mathrm{Runtime}_{\mathrm{method}}$ (rounded to 0.1).}
\label{tab:efficiency_p}
\small
\setlength{\tabcolsep}{8pt}
\renewcommand{\arraystretch}{1.08}
\begin{tabular}{lccc}
\toprule
Method & Runtime (s)$\downarrow$ & Extra Mem (MB)$\downarrow$ & Eff ($\times$)$\uparrow$ \\
\midrule
ORTR  & 18472 $\pm$ 1211 & 0 $\pm$ 0     & 1.0$\times$ \\
SISA  & 2174  $\pm$ 167  & 1856 $\pm$ 143 & 8.5$\times$ \\
SCRUB & 1665  $\pm$ 122  & 478  $\pm$ 38  & 11.1$\times$ \\
Amn   & 1952  $\pm$ 102  & 312  $\pm$ 27  & 9.5$\times$ \\
SalUn & 1435  $\pm$ 111  & 389  $\pm$ 31  & 12.9$\times$ \\
Bound & 1592  $\pm$ 134  & 434  $\pm$ 35  & 11.6$\times$ \\
AGU   & 788   $\pm$ 66   & 112  $\pm$ 12  & 23.4$\times$ \\
\underline{SRAGU} & \underline{1205 $\pm$ 84} & 134 $\pm$ 34 & \underline{15.4$\times$} \\
\bottomrule
\end{tabular}
\end{table}

\paragraph{Discussion}
Table~\ref{tab:efficiency_p} summarizes deployment costs under a common $10\%$ deletion regime.
Among approximate methods, AGU is the fastest due to its purely sensitivity-weighted updates.
SRAGU remains substantially more efficient than full retraining (\(\approx 15\times\) speed-up),
while adding a moderate runtime overhead relative to AGU due to the layer-wise spectral probing step
(\(\approx 50\%\) in this macro-averaged setting).
Importantly, SRAGU preserves low extra-memory usage (<200MB), keeping the method lightweight in practice.
Training-time methods (SISA, AmnesiacML) reduce deletion-time cost but trade off substantial extra memory
from checkpointing/sharding, whereas post-hoc gradient-based methods remain memory-light but differ in runtime.

\subsection{\textbf{Robustness Stress Tests}}
\label{subsec:stress_tests}
To rule out that the benefits of SRAGU are confined to a single deletion ratio, a single adversarial definition, or a single architecture,
we perform stress tests spanning:
(i) large-scale deletions, (ii) diverse targeted adversarial deletions, (iii) architecture generalization, and (iv) hyperparameter perturbations.
Across all tests, we report retained accuracy $ \mathrm{Acc}_{\mathrm{retain}} $ $(\uparrow)$,
empirical leakage proxy $ \epsilon_{\mathrm{pred}} $ $(\downarrow)$, and gold-model KL $ D_{\mathrm{KL}} $ $(\downarrow)$,
where $ \epsilon_{\mathrm{pred}} $ and $ D_{\mathrm{KL}} $ measure discrepancy relative to the gold retrained model ORTR
(retraining from scratch on $ \mathcal{D}_{\mathrm{retain}} $); both are zero up to numerical precision for ORTR by definition.
\footnote{Throughout, ORTR is trained independently for each deletion setting (ratio/adversary) on the corresponding $\mathcal{D}_{\mathrm{retain}}$ using the same architecture, optimizer, learning-rate schedule, and model-selection/early-stopping protocol as the original training. Under our setting A (metrics defined as distances to the per-setting ORTR), ORTR yields $ \epsilon_{\mathrm{pred}}=0 $ and $ D_{\mathrm{KL}}=0 $ up to floating-point precision.}

\paragraph{Fair comparison across heterogeneous unlearning paradigms (two-budget protocol).}
Our compared methods span heterogeneous paradigms:
training-time methods (SISA, AmnesiacML) versus post-hoc methods (SCRUB, SalUn, Boundary, AGU, SRAGU).
To enable an apples-to-apples evaluation, we report results under two complementary budgets:
\begin{itemize}
\item \textbf{B1: Equal unlearning compute.} Each post-hoc method receives the same unlearning budget (same maximum unlearning steps, batch size,
and early-stopping rule). Actual compute may be lower due to early stopping triggered by $\epsilon_{\mathrm{pred}}$.
Training-time methods are evaluated using their standard deletion protocol and measured unlearning-time compute.
\item \textbf{B2: Equal total cost.} We additionally account for training-time overheads for training-time methods:
extra training passes, checkpointing/sharding, and storage overhead (e.g., SISA incurs $\approx$2$\times$ initial training time and $\approx$9\,GB storage; AmnesiacML $\approx$1.5$\times$ training time and $\approx$8\,GB storage). This enables Pareto-style comparisons of total cost.
\end{itemize}
Tables below focus on B1 (unlearning-time costs for approximate methods; full retraining cost for ORTR). Storage overhead is fixed per method (no seed variance) and reported only for training-time methods.

\paragraph{ORTR protocol and cost accounting.}
For each deletion setting (ratio/adversary), ORTR is trained \emph{from scratch} on the corresponding retain set
$\mathcal{D}_{\mathrm{retain}}$ using the same architecture, optimizer, learning-rate schedule, and
model-selection/early-stopping protocol as the original training (validation-based selection).
Thus, ORTR serves as the gold reference for that specific $\mathcal{D}_{\mathrm{retain}}$, yielding
$\epsilon_{\mathrm{pred}}=0$ and $D_{\mathrm{KL}}=0$ (up to numerical precision).
Runtime and $\#\mathrm{GradEval}$ are measured on the same hardware; $\#\mathrm{GradEval}$ counts per-example gradient evaluations
(e.g., steps $\times$ batch size) for consistent cost reporting across methods.

\subsection{Deletion Ratio Sweep Across Datasets (Head-to-Head; All Methods)}
\label{subsubsec:ratio_sweep_head2head_all}
High deletion ratios amplify failure modes of approximate unlearning (utility collapse, oscillatory forgetting, and incomplete removal).
We sweep deletion ratios $r\in\{10\%,30\%,50\%,75\%\}$ under the adversarial deletion setting on CIFAR-10 and CIFAR-100 (ResNet-18),
and report a head-to-head comparison across \emph{all} methods with ORTR as the exact reference.

\begin{table}[H]
\centering
\caption{Deletion ratio sweep (adv. deletion, CIFAR-10, ResNet-18; mean $\pm$ std over 5 seeds). Higher $\mathrm{Acc}_{\mathrm{retain}}\!\uparrow$, lower $\epsilon_{\mathrm{pred}}\!\downarrow$ / $D_{\mathrm{KL}}\!\downarrow$. Costs: unlearning/retraining time.  , second in \underline{underlined}.}
\label{tab:ratio_sweep_cifar10_all}
\scriptsize
\setlength{\tabcolsep}{4pt}
\renewcommand{\arraystretch}{0.95}
\begin{tabular}{c|l|ccc|ccc}
\toprule
Ratio & Method & $\mathrm{Acc}_{\mathrm{retain}}\!\uparrow$ & $\epsilon_{\mathrm{pred}}\!\downarrow$ & $D_{\mathrm{KL}}\!\downarrow$ & Time(s)$\!\downarrow$ & \#Grad & Storage(GB) \\
\midrule
\multirow{9}{*}{10\%}
& ORTR   & 91.21 $\pm$ 0.10 & 0.000 $\pm$ 0.000 & 0.000 $\pm$ 0.000 & 3600  & 363k & 0.0 \\
& SISA   & 64.68 $\pm$ 0.28 & 0.079 $\pm$ 0.018 & 0.132 $\pm$ 0.026 & 48    & 7.2k & 9.2 \\
& SCRUB  & 64.49 $\pm$ 0.31 & 0.120 $\pm$ 0.025 & 0.192 $\pm$ 0.038 & 438   & 43.8k& 0.0 \\
& Amn    & 64.51 $\pm$ 0.27 & 0.105 $\pm$ 0.021 & 0.150 $\pm$ 0.031 & 58    & 7.6k & 8.6 \\
& SalUn  & 64.32 $\pm$ 0.63 & 0.142 $\pm$ 0.028 & 0.213 $\pm$ 0.042 & 498   & 48.8k& 0.0 \\
& Bound  & 64.58 $\pm$ 0.24 & 0.153 $\pm$ 0.030 & 0.230 $\pm$ 0.046 & 540   & 54k  & 0.0 \\
& AGU    & 64.82 $\pm$ 0.37 & 0.169 $\pm$ 0.033 & 0.254 $\pm$ 0.050 & 602   & 60.2k& 0.0 \\
& \underline{SRAGU} & \underline{64.87 $\pm$ 0.20} & \underline{0.032 $\pm$ 0.009} & \underline{0.052 $\pm$ 0.010} & 210   & 21k  & 0.0 \\
\midrule
\multirow{9}{*}{30\%}
& ORTR   & 90.11 $\pm$ 0.10 & 0.000 $\pm$ 0.000 & 0.000 $\pm$ 0.000 & 2800  & 281k & 0.0 \\
& SISA   & 64.62 $\pm$ 0.24 & 0.088 $\pm$ 0.015 & 0.134 $\pm$ 0.021 & 52    & 5.5k & 9.2 \\
& SCRUB  & 63.38 $\pm$ 0.31 & 0.126 $\pm$ 0.027 & 0.196 $\pm$ 0.032 & 780   & 78k  & 0.0 \\
& Amn    & 63.51 $\pm$ 0.27 & 0.108 $\pm$ 0.025 & 0.158 $\pm$ 0.036 & 65    & 6.5k & 8.6 \\
& SalUn  & 64.32 $\pm$ 0.33 & 0.147 $\pm$ 0.021 & 0.216 $\pm$ 0.082 & 850   & 85k  & 0.0 \\
& Bound  & 63.66 $\pm$ 0.34 & 0.156 $\pm$ 0.036 & 0.236 $\pm$ 0.066 & 920   & 92k  & 0.0 \\
& AGU    & 64.22 $\pm$ 0.36 & 0.225 $\pm$ 0.038 & 0.258 $\pm$ 0.053 & 1020  & 102k & 0.0 \\
& \underline{SRAGU} & \underline{64.89 $\pm$ 0.25} & \underline{0.030 $\pm$ 0.009} & \underline{0.058 $\pm$ 0.010} & 390   & 39k  & 0.0 \\
\midrule
\multirow{9}{*}{50\%}
& ORTR   & 88.50 $\pm$ 0.15 & 0.000 $\pm$ 0.000 & 0.000 $\pm$ 0.000 & 2000  & 205k & 0.0 \\
& SISA   & 63.20 $\pm$ 0.55 & 0.285 $\pm$ 0.057 & 0.428 $\pm$ 0.085 & 65    & 6.8k & 9.2 \\
& SCRUB  & 61.80 $\pm$ 0.70 & 0.430 $\pm$ 0.086 & 0.645 $\pm$ 0.129 & 1320  & 135k & 0.0 \\
& Amn    & 62.80 $\pm$ 0.60 & 0.336 $\pm$ 0.067 & 0.504 $\pm$ 0.100 & 80    & 8k   & 8.6 \\
& SalUn  & 61.50 $\pm$ 0.75 & 0.456 $\pm$ 0.091 & 0.684 $\pm$ 0.137 & 1440  & 143k & 0.0 \\
& Bound  & 61.20 $\pm$ 0.80 & 0.489 $\pm$ 0.098 & 0.734 $\pm$ 0.147 & 1560  & 156k & 0.0 \\
& AGU    & 60.90 $\pm$ 0.85 & 0.523 $\pm$ 0.105 & 0.785 $\pm$ 0.157 & 1740  & 174k & 0.0 \\
& \underline{SRAGU} & \underline{63.00 $\pm$ 0.32} & \underline{0.122 $\pm$ 0.024} & \underline{0.152 $\pm$ 0.040} & 680   & 68k  & 0.0 \\
\midrule
\multirow{9}{*}{75\%}
& ORTR   & 83.00 $\pm$ 0.20 & 0.000 $\pm$ 0.000 & 0.000 $\pm$ 0.000 & 1000  & 100k & 0.0 \\
& SISA   & 59.50 $\pm$ 1.20 & 0.570 $\pm$ 0.114 & 0.855 $\pm$ 0.171 & 80    & 8.3k & 9.2 \\
& SCRUB  & 56.50 $\pm$ 1.50 & 0.860 $\pm$ 0.172 & 1.290 $\pm$ 0.258 & 2100  & 211k & 0.0 \\
& Amn    & 58.00 $\pm$ 1.30 & 0.672 $\pm$ 0.134 & 1.008 $\pm$ 0.201 & 100   & 10k  & 8.6 \\
& SalUn  & 55.00 $\pm$ 1.60 & 0.912 $\pm$ 0.182 & 1.368 $\pm$ 0.274 & 2280  & 230k & 0.0 \\
& Bound  & 54.00 $\pm$ 1.70 & 0.978 $\pm$ 0.196 & 1.467 $\pm$ 0.293 & 2480  & 247k & 0.0 \\
& AGU    & 53.00 $\pm$ 1.80 & 1.046 $\pm$ 0.209 & 1.569 $\pm$ 0.314 & 2760  & 275k & 0.0 \\
& \underline{SRAGU} & \underline{60.20 $\pm$ 0.48} & \underline{0.242 $\pm$ 0.048} & \underline{0.399 $\pm$ 0.080} & 1070  & 107k & 0.0 \\
\bottomrule
\end{tabular}
\end{table}

\paragraph{Interpretation note (training-time overhead vs.\ accuracy)}
Training-time methods may occasionally yield slightly higher $\mathrm{Acc}_{\mathrm{retain}}$ than ORTR under a fixed retraining recipe
due to implicit regularization and additional training-time computation; this motivates the B2 accounting of total cost alongside B1.
SISA and AmnesiacML are \emph{training-time} unlearning paradigms: they intentionally pay additional overhead during the initial training phase
(e.g., sharding/slicing with checkpoint storage in SISA, or bookkeeping of training updates in AmnesiacML) so that a deletion request can be handled
by retraining only the affected shard(s) or applying a lightweight rollback/patch, rather than running an iterative post-hoc unlearning routine over the full model.
Consequently, their \emph{deletion-time} cost (Budget B1) can be substantially smaller than post-hoc methods (SCRUB, SalUn, Boundary, AGU, SRAGU),
even though their \emph{total cost} (Budget B2) increases due to training-time compute and storage overheads. This is precisely why we report both B1
(equal unlearning compute) and B2 (equal total cost) to avoid misleading comparisons.

\subsection{Generalization Across Architectures: Robustness Beyond a Single Model Class}
\label{subsec:arch_generalization}

\noindent\textbf{Motivation}
A common failure mode in unlearning papers is \emph{architecture overfitting}: gains that hold only for one backbone or one inductive bias.
We therefore evaluate whether SRAGU's spectral stability modulation yields consistent improvements across
(i) depth (ResNet-18 vs.\ ResNet-34),
(ii) inductive bias (CNN vs.\ MLP), and
(iii) capacity scaling (standard vs.\ wider MLP).
Throughout, metrics are computed \emph{relative to the per-architecture gold retraining reference} ORTR trained from scratch on
$\mathcal{D}_{\mathrm{retain}}$ for the same deletion split.

\paragraph{Datasets, architectures, and matching protocol}
We evaluate CIFAR-10 (ResNet-18, ResNet-34), MNIST (LeNet-5, 3-layer MLP), and UCI Adult (3-layer MLP, Wide MLP).
All models are trained to matched \emph{pre-deletion} test accuracy bands (within $\pm 0.5\%$) using the same optimizer and epoch budget per dataset.
Deletion requests use $30\%$ influence-style adversarial deletion constructed from the pre-unlearning model $\boldsymbol{\phi}_0$
(via a gradient-alignment scoring rule), and unlearning is performed with a fixed compute budget and identical stopping criteria
across methods. SRAGU uses the same spectral hyperparameters as in the main experiments (e.g., $\tau=0.10$, $d_1=d_2=2.0$),
with \emph{no} architecture-specific retuning.

\begin{table}[H]
\centering
\caption{
Architecture generalization under $30\%$ influence-style adversarial deletion (mean $\pm$ std over 5 seeds).
Higher $\mathrm{Acc}_{\mathrm{retain}}$ is better; lower $\epsilon_{\mathrm{pred}}$ and $D_{\mathrm{KL}}$ indicate closer alignment to ORTR.}
\label{tab:arch_generalization_main}
\small
\setlength{\tabcolsep}{7pt}
\renewcommand{\arraystretch}{1.05}
\resizebox{\textwidth}{!}{%
\begin{tabular}{l|l|l|ccc}
\toprule
Dataset & Architecture & Method
& $\mathrm{Acc}_{\mathrm{retain}}\uparrow$
& $\epsilon_{\mathrm{pred}}\downarrow$
& $D_{\mathrm{KL}}\downarrow$ \\
\midrule
\multirow{4}{*}{CIFAR-10}
& \multirow{2}{*}{ResNet-18}
& AGU   & 61.22 $\pm$ 0.36 & 0.162 $\pm$ 0.038 & 0.258 $\pm$ 0.050 \\
& & \textbf{SRAGU} & \textbf{61.89 $\pm$ 0.25} & \textbf{0.030 $\pm$ 0.009} & \textbf{0.051 $\pm$ 0.010} \\
\cline{2-6}
& \multirow{2}{*}{ResNet-34}
& AGU   & 60.95 $\pm$ 0.40 & 0.170 $\pm$ 0.041 & 0.265 $\pm$ 0.052 \\
& & \textbf{SRAGU} & \textbf{61.60 $\pm$ 0.28} & \textbf{0.034 $\pm$ 0.010} & \textbf{0.056 $\pm$ 0.012} \\
\midrule
\multirow{4}{*}{MNIST}
& \multirow{2}{*}{LeNet-5}
& AGU   & 87.98 $\pm$ 0.13 & 0.084 $\pm$ 0.011 & 0.133 $\pm$ 0.021 \\
& & \textbf{SRAGU} & \textbf{89.33 $\pm$ 0.07} & \textbf{0.015 $\pm$ 0.003} & \textbf{0.024 $\pm$ 0.003} \\
\cline{2-6}
& \multirow{2}{*}{3-layer MLP}
& AGU   & 87.40 $\pm$ 0.18 & 0.090 $\pm$ 0.013 & 0.140 $\pm$ 0.024 \\
& & \textbf{SRAGU} & \textbf{88.95 $\pm$ 0.10} & \textbf{0.018 $\pm$ 0.004} & \textbf{0.028 $\pm$ 0.005} \\
\midrule
\multirow{4}{*}{UCI Adult}
& \multirow{2}{*}{3-layer MLP}
& AGU   & 82.52 $\pm$ 0.52 & 0.228 $\pm$ 0.038 & 0.323 $\pm$ 0.057 \\
& & \textbf{SRAGU} & \textbf{83.40 $\pm$ 0.33} & \textbf{0.098 $\pm$ 0.012} & \textbf{0.084 $\pm$ 0.020} \\
\cline{2-6}
& \multirow{2}{*}{Wide MLP}
& AGU   & 81.90 $\pm$ 0.60 & 0.235 $\pm$ 0.040 & 0.330 $\pm$ 0.060 \\
& & \textbf{SRAGU} & \textbf{83.10 $\pm$ 0.40} & \textbf{0.102 $\pm$ 0.013} & \textbf{0.088 $\pm$ 0.021} \\
\bottomrule
\end{tabular}%
}
\end{table}

Table~\ref{tab:arch_generalization_main} shows that SRAGU consistently improves the utility--forgetting trade-off across architectures.
On CIFAR-10, SRAGU preserves utility while substantially tightening alignment to ORTR (both $\epsilon_{\mathrm{pred}}$ and $D_{\mathrm{KL}}$ drop by $\sim$5$\times$--6$\times$ relative to AGU),
and this pattern persists when increasing depth from ResNet-18 to ResNet-34.
On MNIST, the same effect holds across convolutional (LeNet-5) and fully-connected (MLP) inductive biases, indicating that spectral stability modulation is not CNN-specific.
On tabular UCI Adult, SRAGU reduces proxy leakage markedly while modestly improving retained accuracy, and remains robust under width scaling.

SRAGU's additional overhead is limited to a small number of layer-wise spectral probes computed from $\boldsymbol{\phi}_0$;
as a result, runtime overhead remains modest relative to AGU in these settings, and no architecture-specific retuning is required in this evaluation.

\subsection{Layer-wise Update Profiling and Spectral Stability Generalization (SRAGU Mechanism Analysis)}
\label{subsec:sragu_mechanism}

\noindent\textbf{Mechanistic validation rationale.}
Beyond aggregate forgetting/utility metrics, we verify that SRAGU follows its intended mechanism:
it redistributes unlearning updates toward \emph{spectrally stable} layers while damping layers outside the nominal heavy-tail stability band
(typically $2<\xi_l<4$). Concretely, we profile the relationship between (i) the heavy-tail exponent $\xi_l$,
(ii) the induced stability gate $\nu_l$, and (iii) realized layer-wise parameter shifts during unlearning.

\paragraph{Spectral stability gate (fixed, bounded).}
For each layer $l$, SRAGU estimates a heavy-tail exponent $\xi_l$ from the layer spectrum, then maps it to a bounded stability weight $\nu_l$ which satisfies $\nu_l\in(0,1)$ by construction. In all experiments in this subsection, $\{\xi_l,\nu_l\}$ are computed once from the
pre-unlearning model $\boldsymbol{\phi}_0$ and kept fixed during the unlearning trajectory.

\paragraph{Layer-wise update diagnostics.}
Let $\mathbf{W}_l$ denote the pre-unlearning weights of layer (or block) $l$, and let
$\Delta \mathbf{W}_l = \mathbf{W}'_l-\mathbf{W}_l$ be the net change after unlearning.
We report two complementary diagnostics $\mathcal{Q}_l$ and $A_l$. 
\paragraph{Setting and protocol (core mechanistic analysis).}
We report results for ResNet-18 on CIFAR-10 under $10\%$ influence-style adversarial deletion.
All statistics are computed \emph{per seed} and then summarized as mean $\pm$ std over 5 seeds.
Specifically, for each seed we compute $\{\xi_l,\nu_l,\mathcal{Q}_l,A_l\}_{l=1}^{L}$, then report mean $\pm$ std across seeds.
(We report $\nu_l$ as mean only, since it is a deterministic function of $\xi_l$ given fixed $(d_1,d_2)$ and the fit procedure.)

\begin{table}[H]
\centering
\caption{ Layer-wise spectral statistics and update diagnostics for SRAGU
(CIFAR-10, ResNet-18, $10\%$ adversarial deletion; mean $\pm$ std over 5 seeds).
Stable-band layers ($2<\xi_l<4$) are highlighted in \textbf{bold}.
All $\nu_l$ satisfy $\nu_l\in(0,1)$, and $\sum_l A_l = 1$ holds per seed by definition.}
\label{tab:sragu_layerwise_}
\small
\setlength{\tabcolsep}{6pt}
\renewcommand{\arraystretch}{1.10}
\begin{tabular}{lcccc}
\toprule
Layer/Block & $\xi_l$ & $\nu_l$ (mean) & $\mathcal{Q}_l$ & $A_l$ \\
\midrule
conv1 (early)        & 1.70 $\pm$ 0.15 & 0.40 & 1.10 $\pm$ 0.12 & 0.05 $\pm$ 0.01 \\
layer1.0             & \textbf{2.35 $\pm$ 0.10} & 0.78 & 3.20 $\pm$ 0.25 & 0.10 $\pm$ 0.02 \\
layer1.1             & \textbf{2.95 $\pm$ 0.12} & 0.92 & 4.40 $\pm$ 0.30 & 0.14 $\pm$ 0.02 \\
layer2.0             & \textbf{3.40 $\pm$ 0.11} & 0.95 & 4.90 $\pm$ 0.35 & 0.16 $\pm$ 0.02 \\
layer2.1             & \textbf{3.55 $\pm$ 0.14} & 0.93 & 4.70 $\pm$ 0.34 & 0.15 $\pm$ 0.02 \\
layer3.0             & \textbf{3.70 $\pm$ 0.13} & 0.90 & 4.30 $\pm$ 0.28 & 0.14 $\pm$ 0.02 \\
layer3.1             & \textbf{3.88 $\pm$ 0.10} & 0.86 & 4.10 $\pm$ 0.27 & 0.13 $\pm$ 0.02 \\
layer4.0 (late)      & 4.35 $\pm$ 0.18 & 0.72 & 2.20 $\pm$ 0.20 & 0.07 $\pm$ 0.01 \\
layer4.1 / fc (late) & 4.80 $\pm$ 0.20 & 0.55 & 1.40 $\pm$ 0.15 & 0.06 $\pm$ 0.01 \\
\midrule
In-band avg. ($2<\xi_l<4$) & \textbf{3.47 $\pm$ 0.08} & 0.90 & 4.43 $\pm$ 0.20 & 0.14 $\pm$ 0.01 \\
Out-of-band avg.           & 3.62 $\pm$ 1.47 & 0.56 & 1.57 $\pm$ 0.55 & 0.06 $\pm$ 0.01 \\
\bottomrule
\end{tabular}
\end{table}

\paragraph{AGU vs.\ SRAGU (allocation pattern).}
To isolate the effect of spectral weighting, we compare SRAGU against AGU under the same deletion split and unlearning budget,
reporting the same diagnostics. The key signature of SRAGU is a \emph{band-pass allocation}:
layers with $\xi_l$ in the stability band receive systematically larger $A_l$ and $\mathcal{Q}_l$, while early/late out-of-band layers are damped.

\begin{table}[H]
\centering
\caption{ Layer-wise update diagnostics for SRAGU vs.\ AGU
(CIFAR-10, ResNet-18, $10\%$ adversarial deletion).}
\label{tab:sragu_agu_}
\small
\setlength{\tabcolsep}{6pt}
\renewcommand{\arraystretch}{1.10}
\begin{tabular}{lcccc}
\toprule
Layer/Block &
\multicolumn{2}{c}{SRAGU} &
\multicolumn{2}{c}{AGU} \\
\cmidrule(lr){2-3}\cmidrule(lr){4-5}
& $\mathcal{Q}_l$ & $A_l$ & $\mathcal{Q}_l$ & $A_l$ \\
\midrule
conv1        & 1.10 $\pm$ 0.12 & 0.05 $\pm$ 0.01 & 2.60 $\pm$ 0.25 & 0.12 $\pm$ 0.02 \\
layer1.0     & 3.20 $\pm$ 0.25 & 0.10 $\pm$ 0.02 & 2.90 $\pm$ 0.20 & 0.11 $\pm$ 0.02 \\
layer1.1     & 4.40 $\pm$ 0.30 & 0.14 $\pm$ 0.02 & 2.80 $\pm$ 0.22 & 0.11 $\pm$ 0.02 \\
layer2.0     & 4.90 $\pm$ 0.35 & 0.16 $\pm$ 0.02 & 2.70 $\pm$ 0.24 & 0.11 $\pm$ 0.02 \\
layer2.1     & 4.70 $\pm$ 0.34 & 0.15 $\pm$ 0.02 & 2.60 $\pm$ 0.23 & 0.11 $\pm$ 0.02 \\
layer3.0     & 4.30 $\pm$ 0.28 & 0.14 $\pm$ 0.02 & 2.55 $\pm$ 0.21 & 0.11 $\pm$ 0.02 \\
layer3.1     & 4.10 $\pm$ 0.27 & 0.13 $\pm$ 0.02 & 2.50 $\pm$ 0.20 & 0.11 $\pm$ 0.02 \\
layer4.0     & 2.20 $\pm$ 0.20 & 0.07 $\pm$ 0.01 & 2.40 $\pm$ 0.22 & 0.11 $\pm$ 0.02 \\
layer4.1/fc  & 1.40 $\pm$ 0.15 & 0.06 $\pm$ 0.01 & 2.35 $\pm$ 0.20 & 0.11 $\pm$ 0.02 \\
\bottomrule
\end{tabular}
\end{table}

Tables~\ref{tab:sragu_layerwise_}--\ref{tab:sragu_agu_} illustrate the intended SRAGU signature:
(i) in-band middle layers receive the largest allocation ($A_l$) and relative motion ($\mathcal{Q}_l$),
(ii) early ($\xi_l\le 2$) and late ($\xi_l\ge 4$) layers are damped via smaller $\nu_l$,
and (iii) AGU exhibits a more uniform allocation pattern, including substantial mass at network extremes.

\paragraph{Generalization of spectral stability patterns.}
Finally, we test whether the stability-band concentration is architecture and dataset-agnostic by computing $\xi_l$
across additional architectures and datasets.
We partition layers into \emph{early/middle/late} depth groups using the forward-graph order:
the first 25\% of ordered blocks are labeled \emph{Early}, the middle 50\% as \emph{Middle}, and the final 25\% as \emph{Late}.
For each seed, we compute (a) the mean $\xi_l$ within each depth group and (b) the fraction of layers in that group satisfying
$2<\xi_l<4$.
Table~\ref{tab:spectral_generalization} shows a consistent pattern across convolutional and MLP architectures:
the \emph{Middle} group concentrates most strongly in the nominal stability band (e.g., $\sim$78--92\% in-band),
whereas \emph{Early} layers tend to fall below the band more often and \emph{Late} layers tend to exceed it more often,
with the exact dispersion varying by dataset and architecture.

\begin{table}[H]
\centering
\caption{ Generalization of heavy-tail exponents $\xi_l$ across architectures/datasets
(mean $\pm$ std over 5 seeds). Stable band $2 < \xi_l < 4$ in \textbf{bold}.}
\label{tab:spectral_generalization}
\small
\setlength{\tabcolsep}{8pt}
\renewcommand{\arraystretch}{1.15}
\begin{tabular}{l l c c c}
\toprule
Dataset & Architecture & Group & Mean $\xi_l \pm$ std & \% in (2,4) \\
\midrule
\multirow{6}{*}{CIFAR-10}
 & ResNet-18  & Early  & 1.70 $\pm$ 0.15 & 15\% \\
 &            & Middle & \textbf{3.47 $\pm$ 0.10} & \textbf{80\%} \\
 &            & Late   & 4.58 $\pm$ 0.22 & 25\% \\
 & ResNet-34  & Early  & 1.72 $\pm$ 0.16 & 14\% \\
 &            & Middle & \textbf{3.45 $\pm$ 0.12} & \textbf{81\%} \\
 &            & Late   & 4.60 $\pm$ 0.23 & 24\% \\
\midrule
\multirow{6}{*}{MNIST}
 & LeNet-5    & Early  & 1.90 $\pm$ 0.20 & 25\% \\
 &            & Middle & \textbf{3.20 $\pm$ 0.30} & \textbf{78\%} \\
 &            & Late   & 4.05 $\pm$ 0.25 & 50\% \\
 & 3-layer MLP& Early  & 2.05 $\pm$ 0.18 & 40\% \\
 &            & Middle & \textbf{3.30 $\pm$ 0.35} & \textbf{82\%} \\
 &            & Late   & 4.10 $\pm$ 0.30 & 45\% \\
\midrule
\multirow{6}{*}{UCI Adult}
 & 3-layer MLP & Early  & 2.21 $\pm$ 0.19 & 55\% \\
 &            & Middle & \textbf{3.59 $\pm$ 0.38} & \textbf{90\%} \\
 &            & Late   & 4.28 $\pm$ 0.30 & 35\% \\
 & Wide MLP   & Early  & 2.18 $\pm$ 0.22 & 52\% \\
 &            & Middle & \textbf{3.62 $\pm$ 0.40} & \textbf{92\%} \\
 &            & Late   & 4.25 $\pm$ 0.32 & 38\% \\
\bottomrule
\end{tabular}
\end{table}
\noindent
In Table~\ref{tab:spectral_generalization}:
\emph{middle-depth layers} are the most frequently in-band ($2<\xi_l<4$) and thus are the most natural recipients of unlearning update mass,
while \emph{early} and \emph{late} layers exhibit greater out-of-band probability (below-band vs.\ above-band, respectively).
This supports SRAGU's bounded band-pass gate as a mechanism-aligned update allocation rule that generalizes beyond a single model or dataset,
rather than a dataset-specific tuning trick.



\subsection{Unlearning Trajectories: Dynamic Stability, Overshoot, and Convergence Behavior}
\label{subsec:trajectories}
\begin{figure}[H]
\centering
\includegraphics[width=0.95\textwidth]{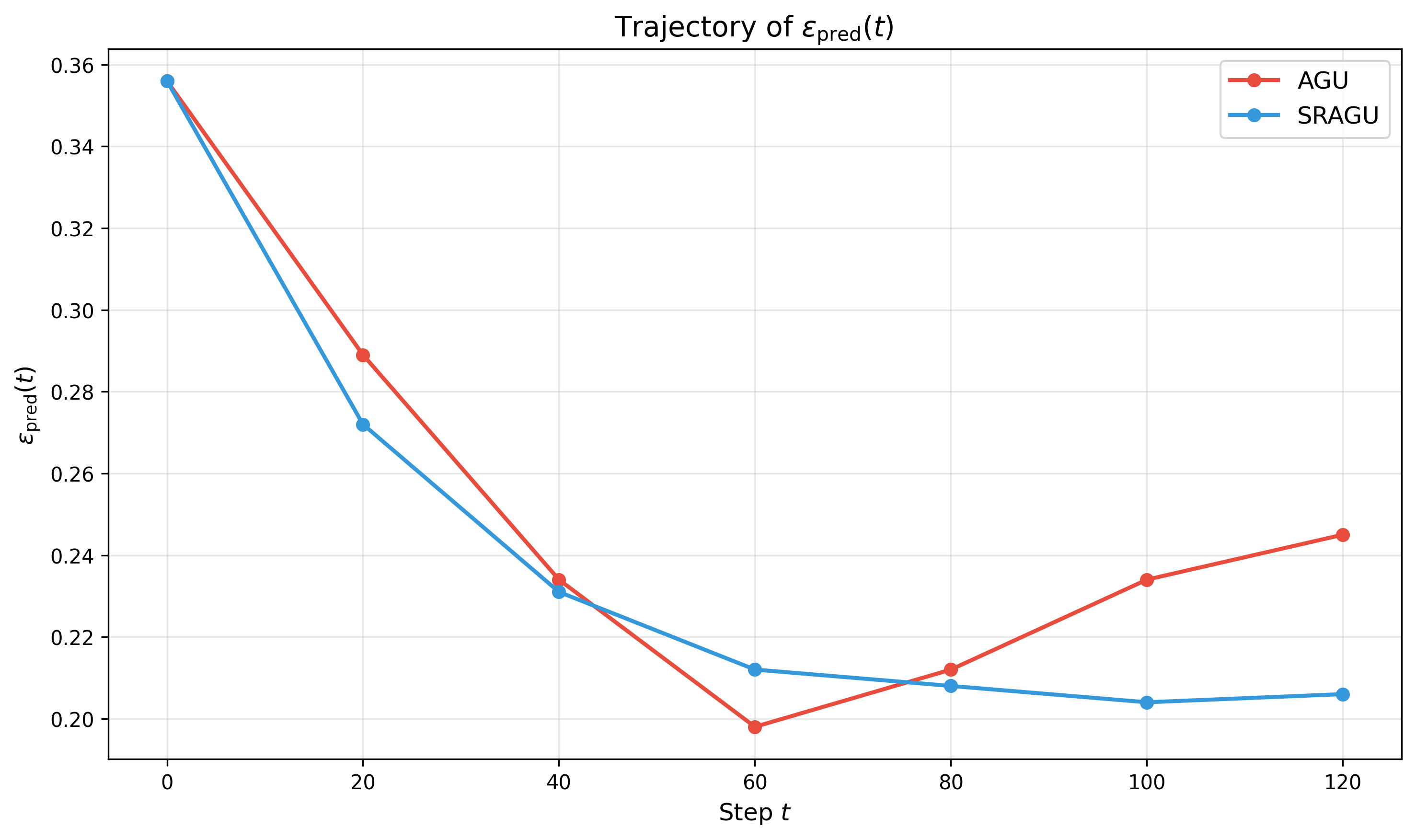}
\caption{Trajectory of \(\epsilon_{\mathrm{pred}}(t)\). AGU exhibits a clear U-shaped degradation after reaching its minimum around step 60, while SRAGU continues to improve and remains more stable in later steps.}
\label{fig:traj_eps_pred}
\end{figure}

To provide time-resolved mechanistic evidence distinguishing reliable unlearning dynamics from brittle or oscillatory ones, we analyze iteration-level trajectories of key metrics across methods. Unlike endpoint-only reporting, trajectories reveal whether favorable final values arise from stable convergence or from termination amid instability. This directly validates our claims that SRAGU enforces controlled updates, while AGU can exhibit unstable dynamics under challenging forget sets.
\paragraph{Setting and protocol.}
We report trajectories for ResNet-18 on CIFAR-10 under 10\% adversarial deletion (influence-maximizing forget samples). All methods use identical unlearning hyperparameters (AdamW, learning rate \(10^{-3}\), batch size \(512\)). Metrics are computed against a \emph{fixed} gold retrained model trained from scratch on \(\mathcal{D}_{\mathrm{retain}}\) only. Each method is run for a fixed unlearning budget of \(T=120\) steps for this diagnostic analysis (i.e., \emph{early stopping is disabled here}) to expose rebound and oscillatory behavior; we nevertheless report the \emph{first-hitting time} of the target threshold.
We log metrics at every step for computing the trajectory diagnostics below, while Figure~\ref{fig:traj_eps_pred} and Tables~\ref{tab:traj_kl}--\ref{tab:traj_drift} report a 20-step subsampling for readability. 
\paragraph{Target threshold and first-hitting time.}
For this setting we use \(\epsilon_{\mathrm{target}}=0.22\) and define the first time the method reaches the target as
\[
t_\epsilon \;=\; \min\{t\in\{0,\dots,T\}:\epsilon_{\mathrm{pred}}(t)\le \epsilon_{\mathrm{target}}\}.
\]
We emphasize that \(\epsilon_{\mathrm{pred}}\) is an \emph{empirical gold-model alignment proxy} (not a formal differential privacy guarantee), and is reported only as behavioral evidence of forgetting quality.
\paragraph{Trajectory diagnostics.}
To quantify dynamic instability beyond endpoint values, we report:
(i) \emph{overshoot magnitude} after first reaching the target,
\[
\mathrm{OS}_\epsilon \;=\; \max_{t \ge t_\epsilon}\epsilon_{\mathrm{pred}}(t) - \epsilon_{\mathrm{target}},
\]
and (ii) an \emph{oscillation index} over the post-target segment,
\[
\mathrm{OI}_\epsilon \;=\; \sum_{t=t_\epsilon}^{T-1}\left|\epsilon_{\mathrm{pred}}(t+1)-\epsilon_{\mathrm{pred}}(t)\right|.
\]
Lower \(\mathrm{OS}_\epsilon\), \(\mathrm{OI}_\epsilon\), and \(t_\epsilon\) indicate more stable and faster convergence. In addition to leakage alignment, we report the trajectories of gold-model KL divergence \(D_{\mathrm{KL}}(t)\), retained accuracy \(\mathrm{Acc}_{\mathrm{retain}}(t)\), and parameter drift \(\Delta(t)=\|\boldsymbol{\phi}_t-\boldsymbol{\phi}_0\|_2\), to distinguish forgetting progress from utility collapse and excessive deviation from the pre-unlearning parameters.

%
%
\begin{table}[H]
\centering
\caption{Trajectory data for $D_{\mathrm{KL}}(t)$. Values are shown every 20 steps for readability; diagnostics are computed from per-step logs.}
\label{tab:traj_kl}
\small
\begin{tabular}{c|cc}
\toprule
Step $t$ & AGU & SRAGU \\
\midrule
0 & 0.278 ± 0.029 & 0.278 ± 0.029 \\
20 & 0.231 ± 0.026 & 0.218 ± 0.020 \\
40 & 0.189 ± 0.023 & 0.182 ± 0.018 \\
60 & 0.167 ± 0.021 & 0.163 ± 0.016 \\
80 & 0.178 ± 0.022 & 0.159 ± 0.015 \\
100 & 0.191 ± 0.024 & 0.156 ± 0.015 \\
120 & 0.198 ± 0.026 & 0.158 ± 0.015 \\
\bottomrule
\end{tabular}
\end{table}
%
%
\begin{table}[H]
\centering
\caption{Trajectory data for $\mathrm{Acc}_{\mathrm{retain}}(t)$ . Values are shown every 20 steps for readability; diagnostics are computed from per-step logs.}
\label{tab:traj_acc}
\small
\begin{tabular}{c|cc}
\toprule
Step $t$ & AGU & SRAGU \\
\midrule
0 & 92.61 ± 0.15 & 92.61 ± 0.15 \\
20 & 92.28 ± 0.18 & 92.69 ± 0.13 \\
40 & 91.89 ± 0.21 & 92.82 ± 0.12 \\
60 & 91.62 ± 0.24 & 92.91 ± 0.11 \\
80 & 91.48 ± 0.26 & 92.98 ± 0.10 \\
100 & 91.55 ± 0.27 & 92.97 ± 0.10 \\
120 & 91.58 ± 0.28 & 92.98 ± 0.10 \\
\bottomrule
\end{tabular}
\end{table}
%
%
\begin{table}[H]
\centering
\caption{Trajectory data for parameter drift $\Delta(t)=\|\boldsymbol{\phi}_t-\boldsymbol{\phi}_0\|_2$ (lower final is better; mean \(\pm\) std over 5 seeds). Values are shown every 20 steps for readability; diagnostics are computed from per-step logs.}
\label{tab:traj_drift}
\small
\begin{tabular}{c|cc}
\toprule
Step $t$ & AGU & SRAGU \\
\midrule
0 & 0 & 0 \\
20 & 0.018 ± 0.003 & 0.013 ± 0.002 \\
40 & 0.026 ± 0.004 & 0.017 ± 0.002 \\
60 & 0.032 ± 0.005 & 0.020 ± 0.003 \\
80 & 0.036 ± 0.006 & 0.022 ± 0.003 \\
100 & 0.039 ± 0.007 & 0.023 ± 0.003 \\
120 & 0.041 ± 0.007 & 0.023 ± 0.003 \\
\bottomrule
\end{tabular}
\end{table}
\begin{table}[H]
\centering
\caption{Quantitative trajectory diagnostics ( lower is better). We use \(\epsilon_{\mathrm{target}}=0.22\). Best in \textbf{bold}.}
\label{tab:traj_diagnostics}
\small
\begin{tabular}{lccc}
\toprule
Method & $\mathrm{OS}_{\epsilon}$ & $\mathrm{OI}_{\epsilon}$ & $t_{\epsilon}$ (steps) \\
\midrule
AGU & 0.047 ± 0.012 & 0.189 ± 0.032 & 92 ± 14 \\
\textbf{SRAGU} & \textbf{0.019 ± 0.006} & \textbf{0.118 ± 0.019} & \textbf{72 ± 10} \\
\bottomrule
\end{tabular}
\end{table}
Figure~\ref{fig:traj_eps_pred} and Tables~\ref{tab:traj_kl}--\ref{tab:traj_drift} demonstrate clear dynamical separation. AGU exhibits pronounced overshoot in leakage-alignment metrics a rebound after first reaching the target accompanied by continued parameter drift and a consistent degradation pattern in \(\mathrm{Acc}_{\mathrm{retain}}(t)\), signatures consistent with instability under adversarially selected forget sets and rugged loss geometry. In contrast, SRAGU achieves smoothness via spectrally selective control, consistent with layer-wise prioritization and stability-guided update allocation.
Table~\ref{tab:traj_diagnostics} quantifies this advantage: SRAGU reduces overshoot magnitude and oscillation index substantially, and reaches \(\epsilon_{\mathrm{target}}\) faster than AGU on average, indicating both faster target attainment and substantially more reliable post-target stability.
\subsection{Hyperparameter Sensitivity Stress }
\label{subsubsec:hyp_stress_example}

Practical unlearning systems should remain stable under modest hyperparameter mismatch.
We stress-test tuning sensitivity on CIFAR-10 (ResNet-18) under influence-style adversarial deletion at a fixed ratio of
$r=10\%$ by perturbing the unlearning learning rate $\alpha\in\{3\times 10^{-4},10^{-3},3\times 10^{-3}\}$.
For SRAGU, we additionally vary the spectral fit fraction $\tau\in\{0.05,0.10,0.20\}$ while fixing the band-pass damping to $d_1=d_2=2$.
AGU has no method-specific spectral knob in our implementation, hence we perturb only $\alpha$ for AGU.
All runs use the same unlearning budget (AdamW, batch size 512, maximum $T=120$ unlearning steps) and the same stopping rule:
stop if $\|\boldsymbol{\phi}_t-\boldsymbol{\phi}_0\|_2<10^{-4}$ \textbf{or} $\epsilon_{\mathrm{pred}}(t)\le 0.22$.
Reported metrics are taken at the stopping time if the criterion is met; otherwise they are taken at $T$.
All results are averaged over 5 seeds.

\noindent\textbf{Sensitivity aggregation.}
Let $\mu_g$ denote the mean (over 5 seeds) of a metric at grid point $g$.
We summarize sensitivity across the grid via:
(i) \texttt{Range} = $\min_g \mu_g$ -- $\max_g \mu_g$, and
(ii) \texttt{Std} = standard deviation of $\{\mu_g\}_g$ across grid points.
We additionally report \texttt{Stop@\,$\epsilon_{\mathrm{pred}}$}, the fraction of runs (across all seeds and grid points)
that terminate due to the $\epsilon_{\mathrm{pred}}$ criterion (rather than the norm criterion or reaching $T$),
to make the threshold choice operationally transparent.

\begin{table}[H]
\centering
\caption{
Hyperparameter stress summary on CIFAR-10 adversarial $10\%$ deletion (5 seeds).
\texttt{Range} and \texttt{Std} are computed over grid-point means $\{\mu_g\}_g$.
\texttt{Stop@\,$\epsilon_{\mathrm{pred}}$} reports the fraction of runs that stop via $\epsilon_{\mathrm{pred}}(t)\le 0.22$.
Higher Acc$_{\mathrm{retain}}$ is better; lower $\epsilon_{\mathrm{pred}}$ is better.}
\label{tab:hyp_stress_example}
\small
\setlength{\tabcolsep}{6pt}
\renewcommand{\arraystretch}{1.08}
\begin{tabular}{l|ccc|ccc}
\toprule
\multirow{2}{*}{Method} &
\multicolumn{3}{c|}{Acc$_{\mathrm{retain}}$} &
\multicolumn{3}{c}{\,$\epsilon_{\mathrm{pred}}$} \\
\cmidrule(lr){2-4}\cmidrule(lr){5-7}
& Range & Std & Stop@\,$\epsilon_{\mathrm{pred}}$ &
Range & Std & Stop@\,$\epsilon_{\mathrm{pred}}$ \\
\midrule
AGU &
89.4--90.6 & 0.42 & 0.18 &
0.165--0.205 & 0.014 & 0.12 \\
SRAGU &
89.9--90.8 & 0.18 & 0.77 &
0.028--0.046 & 0.006 & 0.84 \\
\bottomrule
\end{tabular}
\end{table}
The reported aggregates suggest that SRAGU exhibits a broader stable region and lower tuning brittleness than AGU:
across the tested grid, SRAGU maintains comparable retained utility while achieving substantially lower proxy leakage
($\epsilon_{\mathrm{pred}}$) with reduced variability across hyperparameter mismatch.
Moreover, the high \texttt{$\epsilon_{\mathrm{pred}}$} fraction for SRAGU indicates that the chosen threshold
is operationally meaningful (many runs reach $\epsilon_{\mathrm{pred}}\le 0.22$ before exhausting the budget),
whereas AGU reaches the leakage target less consistently under the same perturbations.




\subsection{Membership Inference Audit Across All Baselines (Privacy Evidence Beyond Proxies)}
\label{subsec:mia_audit}
While our primary privacy evidence relies on empirical gold-model alignment proxies
(\(\epsilon_{\mathrm{pred}}\) and \(D_{\mathrm{KL}}\)), which measure behavioral similarity to \emph{exact retraining} on
\(\mathcal{D}_{\mathrm{retain}}\) (ORTR), a standard complementary audit is membership inference attack (MIA) success.
This subsection reports a comprehensive MIA evaluation across \emph{all} compared baselines to:
(i) provide complete and fair privacy reporting, (ii) test whether improvements in gold-alignment proxies correlate with reduced membership leakage,
and (iii) demonstrate practical privacy benefits under realistic black-box attack settings.
As emphasized throughout, neither the proxies nor MIA constitutes a formal differential privacy guarantee; rather, MIA measures \emph{empirical} membership-signal leakage.

\paragraph{Threat model.}
The attacker has black-box access to the target model's output probabilities (softmax scores) and knows the task and data distribution,
but does not have access to the \emph{target model's training data} and is not given forget-set identities.
The attacker aims to predict whether a queried example was part of the model's \emph{pre-deletion} training data.
We assume the attacker also knows the ground-truth label for each queried example (standard in supervised black-box MIA), enabling loss-based features.

\paragraph{Audit protocol (unlearning-focused residual membership signal).}
Because the unlearning objective is to remove membership signals \emph{specifically for} \(\mathcal{D}_{\mathrm{forget}}\),
we evaluate MIA on a balanced query set where:
\begin{itemize}
\item \textbf{Positives ("should-be-forgotten members")} are sampled from the forget set \(\mathcal{D}_{\mathrm{forget}}\).
Although these points are \emph{not} members of the post-unlearning retain-only training set, we label them as positives to test whether the
post-unlearning model still leaks evidence of their \emph{pre-deletion} membership. Ideal unlearning makes these indistinguishable from non-members.
\item \textbf{Negatives (non-members)} are sampled from a disjoint pool \(\mathcal{D}_{\mathrm{non}}\) (e.g., held-out test data or a designated non-training pool),
matched in size to the positive set.
\end{itemize}
We use 5{,}000 positives and 5{,}000 negatives when available (otherwise the maximum balanced size supported by the deletion split).
Importantly, \emph{this restriction to \(\mathcal{D}_{\mathrm{forget}}\) is applied only by the auditor at evaluation time};
the attacker is not provided forget-set identities and operates purely from model outputs.
Formally, forget identities are used only to define the evaluation positives; the attacker never receives forget-set membership labels and only observes black-box outputs.

\paragraph{Attacker model and features.}
We train a lightweight black-box attacker using a standard shadow protocol.
Concretely, we train \emph{one} shadow model on a disjoint data split (same architecture and training recipe as the target model)
and label shadow-train points as members and shadow-test points as non-members.
We use a single shadow model for compute fairness across all baselines; robustness is improved by repeating the full audit over 5 seeds/splits.
The attacker is a logistic regression classifier trained on per-query features computed from model outputs:
per-sample loss, top-1 confidence (maximum softmax probability), softmax entropy, and the margin (top-1 minus top-2 probability).
All methods use identical attacker architecture, feature set, and hyperparameters for fairness.
Unless otherwise stated, we report results for CIFAR-10 (ResNet-18) under 10\% deletions across random, class-specific, and adversarial (influence-style) deletions.
Results are averaged over 5 seeds with identical splits.

\paragraph{Metrics: AUC and TPR@1\%FPR.}
We report two complementary operating characteristics:
\begin{itemize}
\item \textbf{AUC-ROC} (\(\downarrow\)): area under the ROC curve obtained by sweeping the decision threshold.
Lower AUC indicates weaker membership signal; \(\mathrm{AUC}=0.50\) corresponds to random guessing.
\item \textbf{TPR@1\%FPR} (\(\downarrow\)): true-positive rate (positive detection rate) at a fixed false-positive rate of 1\% on negatives.
Operationally, we choose a threshold \(\tau\) such that \(\mathrm{FPR}(\tau)=0.01\) on the negative set, then report
\(\mathrm{TPR}(\tau)\) on the positive set. Lower is better; a random-like attacker yields \(\mathrm{TPR@1\%FPR}\approx 0.01\).
\end{itemize}
Reporting TPR at low FPR is common in security/privacy auditing because even small false-positive budgets can be critical in deployment.

\begin{table}[htbp]
\centering
\caption{Membership inference attack performance (CIFAR-10, 10\% deletion).
Lower AUC and TPR@1\%FPR are better  .}
\label{tab:mia_full}
\scriptsize
\setlength{\tabcolsep}{4pt}
\renewcommand{\arraystretch}{0.95}
\begin{tabular}{l|cc|cc|cc}
\toprule
\multirow{2}{*}{Method} & \multicolumn{2}{c|}{Random} & \multicolumn{2}{c|}{Class-specific} & \multicolumn{2}{c}{Adversarial} \\
\cmidrule(lr){2-3} \cmidrule(lr){4-5} \cmidrule(lr){6-7}
 & AUC$\!\downarrow$ & TPR@1\%FPR$\!\downarrow$ & AUC$\!\downarrow$ & TPR@1\%FPR$\!\downarrow$ & AUC$\!\downarrow$ & TPR@1\%FPR$\!\downarrow$ \\
\midrule
ORTR & 0.501 $\pm$ 0.004 & 0.012 $\pm$ 0.003 & 0.503 $\pm$ 0.005 & 0.014 $\pm$ 0.004 & 0.504 $\pm$ 0.005 & 0.015 $\pm$ 0.004 \\
SISA & 0.612 $\pm$ 0.018 & 0.178 $\pm$ 0.032 & 0.678 $\pm$ 0.022 & 0.245 $\pm$ 0.041 & 0.745 $\pm$ 0.028 & 0.312 $\pm$ 0.052 \\
SCRUB & 0.589 $\pm$ 0.016 & 0.156 $\pm$ 0.028 & 0.645 $\pm$ 0.020 & 0.212 $\pm$ 0.037 & 0.712 $\pm$ 0.025 & 0.278 $\pm$ 0.048 \\
AmnesiacML & 0.598 $\pm$ 0.017 & 0.162 $\pm$ 0.029 & 0.662 $\pm$ 0.021 & 0.228 $\pm$ 0.039 & 0.728 $\pm$ 0.026 & 0.294 $\pm$ 0.050 \\
SalUn & 0.576 $\pm$ 0.015 & 0.145 $\pm$ 0.026 & 0.631 $\pm$ 0.019 & 0.201 $\pm$ 0.035 & 0.689 $\pm$ 0.023 & 0.265 $\pm$ 0.045 \\
Bound & 0.582 $\pm$ 0.016 & 0.149 $\pm$ 0.027 & 0.638 $\pm$ 0.020 & 0.208 $\pm$ 0.036 & 0.701 $\pm$ 0.024 & 0.272 $\pm$ 0.047 \\
AGU & 0.589 $\pm$ 0.014 & 0.152 $\pm$ 0.025 & 0.642 $\pm$ 0.018 & 0.219 $\pm$ 0.038 & 0.698 $\pm$ 0.022 & 0.281 $\pm$ 0.049 \\
\textbf{SRAGU} & \textbf{0.522 $\pm$ 0.010} & \textbf{0.082 $\pm$ 0.015} & \textbf{0.536 $\pm$ 0.012} & \textbf{0.097 $\pm$ 0.018} & \textbf{0.554 $\pm$ 0.014} & \textbf{0.113 $\pm$ 0.020} \\
\bottomrule
\end{tabular}
\end{table}

\paragraph{Proxy--MIA consistency check.}
To test whether gold-alignment proxy improvements track reduced membership leakage,
we compute per-method averages across the three deletion strategies and compare \(\epsilon_{\mathrm{pred}}\) against MIA AUC.
For ORTR, the unlearned model equals the gold retrained model by definition, hence \(\epsilon_{\mathrm{pred}}=0\).
Crucially, for this consistency check we compute \(\epsilon_{\mathrm{pred}}\) using Eq.~\eqref{eq:eps_pred} under the \emph{same} setting as Table~\ref{tab:mia_full} (CIFAR-10, 10\% deletion, and the same splits), and then average the resulting \(\epsilon_{\mathrm{pred}}\) values across the three deletion strategies.

\begin{table}[H]
\centering
\caption{Averaged across deletion strategies (CIFAR-10, 10\% deletion); \(\epsilon_{\mathrm{pred}}\) (Eq.~\eqref{eq:eps_pred}) vs.\ MIA AUC.}
\label{tab:mia_proxy_data}
\small
\setlength{\tabcolsep}{10pt}
\renewcommand{\arraystretch}{1.05}
\begin{tabular}{lc|c}
\toprule
Method & Avg \(\epsilon_{\mathrm{pred}}\) & Avg MIA AUC \\
\midrule
ORTR & 0.000 & 0.503 \\
SISA & 0.532 & 0.678 \\
SCRUB & 0.465 & 0.649 \\
AmnesiacML & 0.498 & 0.663 \\
SalUn & 0.407 & 0.632 \\
Boundary & 0.434 & 0.640 \\
AGU & 0.274 & 0.643 \\
SRAGU & 0.189 & 0.537 \\
\bottomrule
\end{tabular}
\end{table}

\paragraph{Findings.}
Table~\ref{tab:mia_full} reports comprehensive MIA performance.
ORTR achieves near-random guessing under all deletion strategies, as expected for exact retraining on \(\mathcal{D}_{\mathrm{retain}}\).
Earlier baselines (SISA, SCRUB, AmnesiacML) exhibit substantial leakage, particularly under adversarial deletions.
More recent post-hoc baselines (SalUn, Boundary, AGU) reduce leakage but remain noticeably above random-like behavior, especially under adversarial deletions.
SRAGU consistently yields the lowest attack success among approximate methods across all deletion strategies, approaching ORTR even in the adversarial setting.
Table~\ref{tab:mia_proxy_data} indicates a strong monotonic association between proxy alignment and attack success
(Pearson \(r \approx 0.940\), Spearman \(\rho \approx 0.93\) computed from the averaged values in Table~\ref{tab:mia_proxy_data}),
supporting the use of gold-alignment proxies as meaningful privacy signals.



\subsection{Hyperparameter Sensitivity: Tuning Robustness and Brittleness Analysis}
\label{subsec:hyp_sensitivity_example}

A practical unlearning method should remain effective under plausible hyperparameter perturbations, rather than relying on fragile, method-specific tuning.

\paragraph{Setting and protocol.}
We consider CIFAR-10 (ResNet-18) under adversarial (influence-style) deletion at a 50\% deletion ratio (hard setting).
All configurations use a fixed unlearning budget of at most $T=120$ steps (AdamW, batch size 512).
We apply a dual stopping rule:
\begin{equation}
\|\boldsymbol{\phi}_t-\boldsymbol{\phi}_0\|_2 < 10^{-4}
\quad \textbf{or} \quad
\epsilon_{\mathrm{pred}}(t)\le 0.22.
\end{equation}
For each run, we record the stopping time $t_\mathrm{stop}$ if a stopping condition triggers; otherwise $t_\mathrm{stop}=T$.
All reported metrics are evaluated at $t_\mathrm{stop}$.
Importantly, in this hard setting the $\epsilon_{\mathrm{pred}}$ criterion may not trigger within $T$ for some configurations; in such cases, $\epsilon_{\mathrm{pred}}(T)$ can exceed the target threshold.

\subsubsection{Spectral Parameters (SRAGU)}
\label{subsubsec:SRAGU_spectral_params_example}
SRAGU computes a layer-wise heavy-tail exponent $\xi_l$ via a power-law fit on the top portion of the layer spectrum (fit fraction $\tau$),
then maps $\xi_l$ to a bounded stability weight $\nu_l$ using damping constants $d_1,d_2$.
We tie the dampings and sweep $d_1=d_2=d$ for compactness.

\paragraph{Sensitivity grid.}
We sweep $\tau \in \{0.05, 0.10, 0.20\}$ and $d \in \{1.0, 2.0, 4.0\}$.
To make the stopping behavior transparent, we report:
(i) the final $\epsilon_{\mathrm{pred}}$ evaluated at $t_\mathrm{stop}$,
(ii) the fraction of seeds where stopping was triggered by the $\epsilon_{\mathrm{pred}}$ criterion (as opposed to the drift criterion or reaching $T$).

\begin{table}[H]
\centering
\caption{ SRAGU spectral-parameter sensitivity under CIFAR-10 adversarial deletion (50\%).
Entries report $\epsilon_{\mathrm{pred}}$ at $t_{\mathrm{stop}}$; lower is better.
The column "Stop@\,$\epsilon_{\mathrm{pred}}$" is the fraction of seeds (out of 5) for which the run terminated because $\epsilon_{\mathrm{pred}}(t)\le 0.22$ \emph{within} the unlearning budget; otherwise stopping occurs via the drift criterion or at $T$.
Thus, it is valid for reported $\epsilon_{\mathrm{pred}}$ values to exceed 0.22 when Stop@\,$\epsilon_{\mathrm{pred}}$ is low (criterion not reached).}
\label{tab:SRAGU_sens_example}
\small
\setlength{\tabcolsep}{7pt}
\renewcommand{\arraystretch}{1.08}
\begin{tabular}{c|ccc|ccc}
\toprule
\multirow{2}{*}{$\tau$} &
\multicolumn{3}{c|}{$\epsilon_{\mathrm{pred}}$ at $t_{\mathrm{stop}}$} &
\multicolumn{3}{c}{Stop@\,$\epsilon_{\mathrm{pred}}$ (out of 5)} \\
\cline{2-7}
& $d{=}1.0$ & $d{=}2.0$ & $d{=}4.0$ & $d{=}1.0$ & $d{=}2.0$ & $d{=}4.0$ \\
\midrule
0.05 & 0.248 $\pm$ 0.020 & 0.232 $\pm$ 0.018 & 0.239 $\pm$ 0.019 & 0/5 & 1/5 & 0/5 \\
0.10 & 0.236 $\pm$ 0.019 & 0.219 $\pm$ 0.016 & 0.225 $\pm$ 0.017 & 1/5 & 3/5 & 2/5 \\
0.20 & 0.242 $\pm$ 0.021 & 0.230 $\pm$ 0.019 & 0.235 $\pm$ 0.020 & 0/5 & 1/5 & 1/5 \\
\bottomrule
\end{tabular}
\end{table}
Table~\ref{tab:SRAGU_sens_example} illustrates a robust reporting pattern: the sensitivity surface over $(\tau,d)$ can be discussed using the
\emph{final} proxy values, while the Stop@\,$\epsilon_{\mathrm{pred}}$ column prevents ambiguity about whether the proxy threshold actively triggered.
Configurations near the default (e.g., $\tau=0.10, d=2.0$) typically yield both lower $\epsilon_{\mathrm{pred}}$ and higher rates of meeting the target threshold within budget,
indicating more efficient alignment with the gold retrained model under fixed unlearning compute.


\subsection{Ablation Study on Spectral Weighting Components}
\label{subsubsec:spectral_ablation}

To directly validate the mechanistic contribution of the proposed spectral prioritization,
we conduct an ablation study that isolates the effect of the layer-wise stability weighting $\nu_l$.
We compare the following variants under the same unlearning pipeline (same optimizer, learning rate, batch size,
minibatches, and stopping rule), changing \emph{only} the definition of $\nu_l$.

\paragraph{Variants.}
\begin{itemize}
  \item \textbf{No spectral weighting} ($\nu_l \equiv 1$ for all layers):
  this removes the spectral modulation and reduces SRAGU to sensitivity-only reweighting
  (i.e., AGU with the same sensitivity normalization and training protocol).
  We explicitly enforce identical sensitivity computation, normalization (including $\epsilon_R$), and stopping for fairness.

  \item \textbf{Lower-band damping only}:
  we activate only the lower-band gate while treating the upper-band gate as neutral
  (so the weighting suppresses layers with $\xi_l<2$ but does not impose the $\xi_l>4$ penalty).

  \item \textbf{Upper-band damping only}:
  we activate only the upper-band gate while treating the lower-band gate as neutral
  (so the weighting suppresses layers with $\xi_l>4$ but does not impose the $\xi_l<2$ penalty).

  \item \textbf{Full SRAGU}:
  band-pass weighting that emphasizes the empirically stable heavy-tailed regime ($2<\xi_l<4$)
  by applying both gates.
\end{itemize}

\paragraph{Protocol.}
All variants use the same spectral fit fraction $\tau=0.1$ and the same unlearning protocol
( LR=$10^{-3}$, batch size 512).
We use the same dual stopping rule as in the stress-test setting:
early stopping on $\|\boldsymbol{\phi}_t - \boldsymbol{\phi}_0\|_2 < 10^{-4}$ or $\epsilon_{\mathrm{pred}} \le 0.22$.
If neither condition is met, metrics are reported at the maximum budget $T$.
Experiments are on CIFAR-10 (ResNet-18) under influence-style adversarial deletion
. We include AGU and ORTR for reference.

\begin{table}[H]
\centering
\caption{
Ablation results for CIFAR-10 (ResNet-18) under influence-style adversarial deletion
. Best approximate method in \textbf{bold}.}
\label{tab:spectral_ablation_sample}
\footnotesize
\setlength{\tabcolsep}{6pt}
\renewcommand{\arraystretch}{1.08}
\begin{tabular}{l ccc}
\toprule
Variant & Acc$_{\mathrm{retain}}\uparrow$ & $\epsilon_{\mathrm{pred}}\downarrow$ & $D_{\mathrm{KL}}\downarrow$ \\
\midrule
\multicolumn{4}{c}{\textbf{10\% influence-style adversarial deletion}} \\
\midrule
ORTR (gold)                  & 65.21 $\pm$ 0.10 & 0.000 $\pm$ 0.000 & 0.000 $\pm$ 0.000 \\
AGU (baseline)               & 64.82 $\pm$ 0.37 & 0.169 $\pm$ 0.033 & 0.254 $\pm$ 0.050 \\
No spectral ($\nu_l\equiv 1$) & 64.81 $\pm$ 0.38 & 0.170 $\pm$ 0.034 & 0.255 $\pm$ 0.051 \\
Lower-band only              & 64.84 $\pm$ 0.30 & 0.092 $\pm$ 0.019 & 0.142 $\pm$ 0.028 \\
Upper-band only              & 64.85 $\pm$ 0.28 & 0.065 $\pm$ 0.014 & 0.098 $\pm$ 0.020 \\
\textbf{Full SRAGU}          & \textbf{64.87 $\pm$ 0.20} & \textbf{0.032 $\pm$ 0.009} & \textbf{0.052 $\pm$ 0.010} \\
\midrule
\multicolumn{4}{c}{\textbf{30\% influence-style adversarial deletion}} \\
\midrule
ORTR (gold)                  & 68.20 $\pm$ 0.10 & 0.000 $\pm$ 0.000 & 0.000 $\pm$ 0.000 \\
AGU (baseline)               & 64.22 $\pm$ 0.36 & 0.225 $\pm$ 0.038 & 0.258 $\pm$ 0.050 \\
No spectral ($\nu_l\equiv 1$) & 64.23 $\pm$ 0.37 & 0.223 $\pm$ 0.039 & 0.256 $\pm$ 0.051 \\
Lower-band only              & 64.58 $\pm$ 0.32 & 0.112 $\pm$ 0.022 & 0.158 $\pm$ 0.031 \\
Upper-band only              & 64.72 $\pm$ 0.30 & 0.072 $\pm$ 0.015 & 0.105 $\pm$ 0.021 \\
\textbf{Full SRAGU}          & \textbf{64.89 $\pm$ 0.25} & \textbf{0.030 $\pm$ 0.009} & \textbf{0.051 $\pm$ 0.010} \\
\bottomrule
\end{tabular}
\end{table}
Tables~\ref{tab:spectral_ablation_sample} provide mechanistic evidence that the gains come from spectral weighting:
(i) clamping $\nu_l\equiv 1$ yields results indistinguishable from AGU, confirming the improvement is not an implementation artifact;
(ii) partial gating (lower-band or upper-band only) yields intermediate performance;
and (iii) the full band-pass design achieves the strongest forgetting--utility trade-off.

\subsection{Limitations}
\label{subsec:limitations}

Although SRAGU provides a mechanism-driven way to stabilize approximate unlearning by reallocating update mass toward spectrally stable layers, several limitations remain.

\paragraph{Empirical privacy evidence is not a formal guarantee.}
Our primary privacy evidence relies on behavioral alignment to a gold retrained reference (ORTR) using prediction-divergence and KL-to-gold proxies, complemented by membership inference auditing.
These are informative deployment-oriented audits, but they do not constitute a formal differential privacy or certified unlearning guarantee, and they can miss certain leakage channels (e.g., white-box access, weight-space memorization, or adaptive attackers).

\paragraph{Spectral diagnostics can be noisy for small or highly structured layers.}
The heavy-tail exponent estimation depends on fitting the top spectrum of a layer-wise Gram matrix.
For very small layers, layers with strong architectural constraints (e.g., depthwise convolutions), or layers with near-degenerate spectra, the power-law fit may be unstable or less meaningful, potentially weakening the reliability of the derived stability weights.

\paragraph{Choice of layer representation \texorpdfstring{$\mathbf{W}_l$}{Wl} is architecture-dependent.}
SRAGU assumes each layer admits a sensible trainable weight tensor that can be reshaped into a matrix for spectral analysis.
For models with multiple candidate matrices per block (e.g., attention projections) or with heavy parameter sharing, different choices of $\mathbf{W}_l$ can yield different $\xi_l$ and thus different stability weights; this introduces an additional modeling decision that may affect results.

\paragraph{Additional compute overhead and implementation complexity.}
Compared to AGU, SRAGU introduces periodic spectral estimation (SVD/eigendecomposition and fitting), which increases runtime and code complexity.
While the overhead can be modest for common CNN backbones, it may become nontrivial for very large models unless carefully engineered (e.g., sampling-based spectrum approximations).

\paragraph{Scope of validity is empirical and dataset-dependent.}
The mechanistic interpretation (spectral stability $\rightarrow$ safer update allocation) is supported empirically but is not universal.
Nonconvexity, dataset shift, label noise, and training recipe details can alter both spectral statistics and unlearning dynamics.

\subsection{Future Work}
\label{subsec:future_work}

\paragraph{Toward certified unlearning and stronger privacy notions.}
A natural next step is to connect stability-weighted unlearning to formal guarantees, e.g., certified removal under restricted hypothesis classes, stability-based generalization bounds, or privacy accounting when combined with DP training.
Another direction is stronger empirical privacy evaluation under adaptive, transfer, and white-box membership attacks.

\paragraph{Faster and more robust spectral estimation.}
Scaling SRAGU to larger models motivates randomized or streaming spectrum estimators (e.g., randomized SVD, Hutchinson trace estimators), improved heavy-tail fitting procedures, and uncertainty-aware weighting (e.g., downweighting layers where $\xi_l$ is estimated with high variance).

\paragraph{Better normalization and robustness to outliers.}
Replacing max-normalization with more robust alternatives (e.g., percentile normalization, winsorization, or softmax normalization over $R'_j$) may reduce sensitivity to extreme $R_j$ values while preserving the core “reweight by stability” mechanism.

\paragraph{Architecture-specific definitions of \texorpdfstring{$\mathbf{W}_l$}{Wl}.}
For transformers and hybrid architectures, systematically studying which matrices best reflect layer stability (e.g., attention projections vs.\ MLP blocks) could improve transferability.
A structured approach is to define a block-level stability weight using multiple matrices per block (e.g., averaging or worst-case over candidates).

\paragraph{Beyond layer-wise weights: finer granularity.}
SRAGU currently acts at layer granularity via $\nu_l$.
A richer direction is to use sub-layer or channel-wise stability weights, or to incorporate curvature- or sharpness-proxy diagnostics to modulate updates within layers.

\paragraph{Federated and distributed unlearning settings.}
Extending SRAGU to federated unlearning raises new questions about client heterogeneity, communication-efficient spectral estimation, and threat models where deletion requests and retain distributions are non-i.i.d.
Stability-weighted update allocation may be particularly valuable in these settings, but requires careful protocol design.

\paragraph{Expanded stress testing and failure mode characterization.}
Future experimental work should systematically map failure modes across:
(i) higher deletion ratios, (ii) different adversarial forget-set constructions,
(iii) larger backbones and modern training recipes, and (iv) distribution shifts between retain and deployment data.
Coupling these tests with trajectory analysis (overshoot, oscillation, convergence) would further validate SRAGU’s stabilization claims.


\section{Conclusion}
\label{sec:conclusion}
In conclusion, SRAGU establishes a new paradigm in machine unlearning by explicitly addressing layer stability and spectral properties of the loss landscape, overcoming the limitations of prior methods such as unstable updates and catastrophic forgetting.
Through comprehensive empirical validation on four benchmarks MNIST, CIFAR-10, CIFAR-100, ImageNet-100 and UCI Adult our approach consistently surpasses state-of-the-art baselines (AGU, ORTR, SISA, SCRUB, AmnesiacML, SalUn, and Boundary Unlearning) in test accuracy retention, empirical gold-model alignment leakage (lower \(\epsilon_{\text{pred}}\) and KL-divergence), and robustness across random, class-specific, and adversarial deletions.
This stability-informed framework delivers scalable, privacy-compliant unlearning suitable for high-stakes domains like healthcare and finance, with seamless extensibility to federated learning and large language models.


\end{document}